%% file: main.tex
\documentclass[10pt,twocolumn,letterpaper]{article}

\usepackage[pagenumbers]{cvpr}      

\input{preamble}

\definecolor{cvprblue}{rgb}{0.21,0.49,0.74}
\usepackage[pagebackref,breaklinks,colorlinks,citecolor=cvprblue]{hyperref}

\def\paperID{72} 
\def\confName{3DV\xspace}
\def\confYear{2026\xspace}

\title{Rethinking Metrics and Diffusion Architecture for 3D Point Cloud Generation}

\author{Matteo Bastico\textsuperscript{1}\footnotemark, David Ryckelynck\textsuperscript{3}
Laurent Cort\'e\textsuperscript{1}, Yannick Tillier\textsuperscript{3}, Etienne Decenci\`ere\textsuperscript{2}\\
Mines Paris, Université PSL\\
\textsuperscript{1}Centre des Matériaux (MAT), UMR7633 CNRS, 91003 Evry, France \\
\textsuperscript{2}Centre de Morphologie Mathématique (CMM), 77300 Fontainebleau, France \\
\textsuperscript{3}Centre de Mise en Forme des Matériaux (CEMEF), UMR7635 CNRS, 06904 Sophia Antipolis, France\\
}

\begin{document}
\maketitle

\renewcommand*{\thefootnote}{\fnsymbol{footnote}}\stepcounter{footnote}%
  \footnotetext[1]{Corresponding author: \href{mailto:matteo.bastico@minesparis.psl.eu}{matteo.bastico@minesparis.psl.eu}}
\setcounter{footnote}{0}

\input{sec/0_abstract}    
\input{sec/1_intro}

\input{sec/2_related_works}
\input{sec/3_methods}

\input{sec/4_experiments}

\input{sec/5_conclusions}
\input{sec/6_ack}
{
    \small
    \bibliographystyle{ieeenat_fullname}
    \bibliography{diffusion}
}

\input{sec/X_suppl}

\end{document}

%% file: preamble.tex
\usepackage[dvipsnames]{xcolor}

\usepackage{comment}
\usepackage{tabularx}
\usepackage{multirow}

\newcolumntype{Y}{>{\centering\arraybackslash}X}

\usepackage[ruled,]{algorithm2e}

\newlength\mylen

\SetKwComment{Comment}{$\triangleright$\ }{}

\DeclareMathOperator*{\argmin}{arg\,min}

%% file: sec/0_abstract.tex
\begin{abstract}
As 3D point clouds become a cornerstone of modern technology, the need for sophisticated generative models and reliable evaluation metrics has grown exponentially. In this work, we first expose that some commonly used metrics for evaluating generated point clouds, particularly those based on Chamfer Distance (CD), lack robustness against defects and fail to capture geometric fidelity and local shape consistency when used as quality indicators. We further show that introducing samples alignment prior to distance calculation and replacing CD with Density-Aware Chamfer Distance (DCD) are simple yet essential steps to ensure the consistency and robustness of point cloud generative model evaluation metrics. While existing metrics primarily focus on directly comparing 3D Euclidean coordinates, we present a novel metric, named Surface Normal Concordance (SNC), which approximates surface similarity by comparing estimated point normals. This new metric, when combined with traditional ones, provides a more comprehensive evaluation of the quality of generated samples. Finally, leveraging recent advancements in transformer-based models for point cloud analysis, such as serialized patch attention , we propose a new architecture for generating high-fidelity 3D structures, the Diffusion Point Transformer. We perform extensive experiments and comparisons on the ShapeNet dataset, showing that our model outperforms previous solutions, particularly in terms of quality of generated point clouds, achieving new state-of-the-art. Code available at \url{https://github.com/matteo-bastico/DiffusionPointTransformer}
\end{abstract}

%% file: sec/1_intro.tex
\begin{figure*}
    \centering
    \includegraphics[width=.89\linewidth]{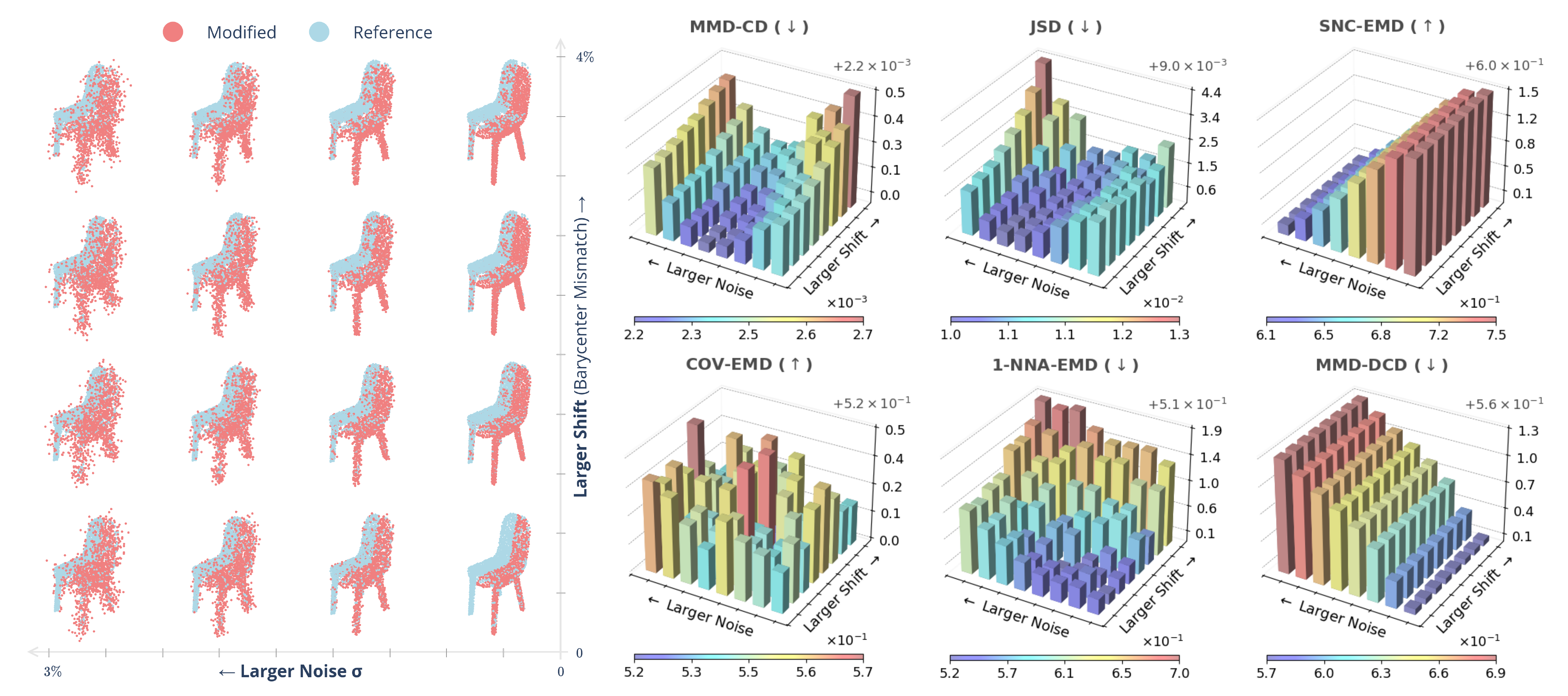}
    \caption{Response of several metrics to random noise and barycenter shift on generated samples. (\textbf{Left}) An example comparing a reference sample (blue) and its modified version (red) as noise and barycenter translations are added in proportion to its diameters. (\textbf{Right}) An overview of the robustness of some traditional metrics (MMD-CD, COV-EMD, JSD and 1-NNA-EMD) and some proposed metrics (SNC-EMD, MMD-DCD) for evaluating point cloud generation.}
    \label{fig:metrics_robustness}
\end{figure*}

\section{Introduction}
\label{sec:intro}
The analysis of 3D point clouds, critical for applications ranging from autonomous vehicles \cite{chen_3d_2021} and robotics \cite{duan_robotics_2021} to the medical domain \cite{yu_3d_2021, liu_grab-net_2023}, faces persistent challenges in collecting and annotating large-scale data. With recent advancements in deep generative models \cite{ruthotto_introduction_2021}, point cloud generation and synthesis have attracted growing interest from the research community, aiming to produce high-fidelity, realistic samples \cite{kim_setvae_2021, gadelha_multiresolution_2018, yang_foldingnet_2018, valsesia_learning_2019, achlioptas_learning_2018, shu_3d_2019, yang_pointflow_2019, klokov_discrete_2020, mo_dit-3d_2023, luo_diffusion_2021, zhou_3d_2021}. Like other generative tasks, this field presents two major challenges: (1) designing effective deep learning architectures and (2) developing robust evaluation methods to ensure fair model comparisons.

Generative AI has achieved significant success across various domains, producing high-quality 2D images \cite{ramesh_hierarchical_2022, peebles_scalable_2023}, among others, mainly leveraging transformer-based architectures \cite{vaswani_attention_2017}. These models are built upon attention mechanisms to capture relationships between input tokens. This makes them inherently suited to point cloud analysis, where understanding spatial relationships between points is essential. As a result, deep learning algorithms for point cloud processing have recently received a significant boost \cite{guo_pct_2021, zhao_point_2021, lai_stratified_2022, wu_point_2022, wu_point_2024, yu_point-bert_2022, wang_serialized_2024, wang_octformer_2023, liu_flatformer_2023}. Classification and segmentation tasks, in particular, have achieved impressive performance thanks to recent developments, such as Point Cloud Transformer (PCT) \cite{guo_pct_2021}, Point Transformer (PT) and its successors \cite{zhao_point_2021, wu_point_2022, wu_point_2024}. Meanwhile, Denoising Diffusion Probabilistic Models (DDPMs) \cite{ho_denoising_2020} have demonstrated immense potential in generative tasks \cite{xia_diffir_2023, huang_fastdiff_2022, ho_video_2022, luo_diffusion_2021, jo_graph_2024, peebles_scalable_2023, dhariwal_diffusion_2021} by employ a forward noising process 
and learning a reverse process that restores the original data. 
Several efforts have been made to apply DDPMs to 3D shapes \cite{luo_diffusion_2021, zhou_3d_2021, zeng_lion_2022, mo_dit-3d_2023, ji_latent_2024, petrov_gem3d_2024}. However, many of these approaches rely on partitioning input data into voxels \cite{mo_dit-3d_2023}, using downsampled encoded tokens \cite{ji_latent_2024}, or leveraging skeletons \cite{petrov_gem3d_2024}, often leading to the loss of local structure details. Despite these advancements, point cloud generation and evaluation remain challenging due to the complexity of 3D data and the difficulty of assessing spatial relationships. As we will show, some traditional point cloud generative model evaluation metrics \cite{yang_pointflow_2019, achlioptas_learning_2018} frequently fail to capture geometric fidelity and structural consistency, especially in the presence of noise and translations on generated samples, slowing progress in developing more robust and reliable solutions. 
Thus, new guidelines for assessment are needed to better meet the demands of real-world applications. In this work, we propose enhancements to existing metrics to improve their stability and better reflect the true quality of generated shapes. Our approach involves performing rigid alignment of synthesized shapes to ensure consistent matching with reference samples, along with incorporating recent improvements of Chamfer Distance (CD) to account for point density rather than relying solely on Euclidean distance, i.e. the Density-aware Chamfer Distance (DCD) \cite{wu_density-aware_2021}. Additionally, we introduce a new metric, the Surface Normal Concordance (SNC), which facilitates the evaluation of point cloud structures by incorporating point normals, particularly in contexts where surface regularity and local geometry are critical for generating realistic synthetic data \cite{ran_surface_2022, huang_surface_2024}. Through a small scale user study, we show that SNC better reflects human visual perception than current quality indicators. 

Furthermore, to enhance the quality of generated point clouds, we introduce a novel plain transformer-based architecture for DDPM, inspired by recent advancements on point cloud processing \cite{zhao_point_2021, wu_point_2022, wu_point_2024}, called Diffusion Point Transformer (DiPT). Unlike existing methods, our model preserves the raw input size (in number of points) throughout its layers, avoiding voxelization or downsampling, which often compromise output surface quality. Experiments on the ShapeNet benchmark \cite{chang_shapenet_2015} show that our point-wise diffusion approach consistently produces higher-fidelity generated samples, demonstrating a clear improvement over previous methods. 

Our contributions are summarized as follows:
\begin{itemize}[noitemsep,topsep=0pt]
    \item We propose new guidelines to improve the evaluation metrics for point cloud generative models. 
    \item We introduce a new metric, Surface Normal Concordance (SNC), to assess the samples quality by also considering point normals rather than only Euclidean distances.
    \item We present Diffusion Point Transformer (DiPT), a novel model for point-wise diffusion that avoids voxelization or downsampling, boosting final quality.
    \item We provide extensive evaluation and comparison of DiPT on the ShapeNet dataset \cite{chang_shapenet_2015} on various object categories.
\end{itemize}

%% file: sec/2_related_works.tex
\section{Related Works}
\label{sec:related_works}

\textbf{Metrics.} Several metrics have been defined to assess the quality of point cloud generative models \cite{yang_pointflow_2019, mo_dit-3d_2023, zhou_3d_2021, achlioptas_learning_2018, triess_realism_2022}.  These metrics always compare a set of generated samples, $S_g$, with a reference set, $S_r$. The Fréchet Point Cloud Distance (FPD) \cite{shu_3d_2019}, inspired by the Fréchet Inception Distance (FID) \cite{heusel_gans_2017}, defined to evaluate 2D image generation, was initially used to measure the distance between real and generated samples in the feature spaces extracted by PointNet \cite{qi_pointnet_2017}. In recent studies, FPD has been replaced by newer metrics that leverage Euclidean distances to quantify point clouds similarity \cite{nguyen_point-set_2021}. Two widely used distance measures for point clouds are the CD and the Earth Mover’s Distance (EMD). CD calculates the sum of the squared Euclidean distances from each point in one point cloud to the nearest point in the other point cloud, while EMD, also known as Wasserstein distance, computes the minimal cost required to transform one point cloud into another. Metrics built on such measures aim to effectively capture both the quality, i.e. realism, of generated samples and/or their diversity or representativeness. Based on these two principles, \textit{Achlioptas et al.} \cite{achlioptas_learning_2018} introduced three key evaluation metrics:
\begin{itemize}
    \item Coverage (\textbf{COV}): Evaluates the diversity of generated samples relative to the reference set.
    \item Minimum Matching Distance (\textbf{MMD}): Measures the average distance to the nearest (i.e., most similar) reference, aiming to capture the quality of generated samples.
    \item Jensen-Shannon Divergence (\textbf{JSD}): Quantifies the similarity between the marginal point distributions of voxelized reference and generated shapes.
\end{itemize}
Recently, to overcome some limitations of these metrics, \textit{Yang et al.} \cite{yang_pointflow_2019} introduced a new metric, the 1-Nearest Neighbour Accuracy (\textbf{1-NNA}) \cite{lopez-paz_revisiting_2017, xu_empirical_2018}. 
It essentially measures to what extent the distributions of $S_g$ and $S_r$ are similar, focusing primarily on the diversity of generated point clouds, with a marginal consideration of quality. 
Furthermore, \textit{Triess et al.} \cite{triess_realism_2022} proposed a learning-based metric to quantify the realism of local regions in LiDAR point clouds. However, this approach requires a proxy classification task trained on both real-world and synthetic point clouds. Following previous works, we refer to a given metric computed with a specific distance measure as \textsc{Metric-Measure} (e.g., MMD-CD refers to MMD calculated using CD). As shown in \cref{fig:metrics_robustness} and discussed in the next section, certain traditional metrics can lead to misleading evaluations. To address this, we introduce metric enhancements, together with SNC, to provide a more reliable and comprehensive assessment of generative models.


Formal definitions of the distance measures and traditional metrics are provided in \cref{sec:metrics_details} of the Supplementary.

\textbf{3D Point Cloud Generation.} Different techniques have been exploited for 3D point cloud generation, mostly deep-learning methods such as Variational AutoEncoders (VAE) \cite{kim_setvae_2021, gadelha_multiresolution_2018, yang_foldingnet_2018}, Generative Adversarial Networks (GANs) \cite{valsesia_learning_2019, achlioptas_learning_2018, shu_3d_2019}, normalized flows \cite{yang_pointflow_2019, klokov_discrete_2020}, and diffusion models \cite{mo_dit-3d_2023, luo_diffusion_2021, zhou_3d_2021}. Among these, FoldingNet \cite{yang_foldingnet_2018} was an early attempt, built upon PointNet \cite{qi_pointnet_2017} to address unsupervised learning challenges on point clouds using a VAE.
SetVAE \cite{kim_setvae_2021} approached point cloud generation as a set generation task using a hierarchical VAE based on a set transformer \cite{lee_set_2019}. ShapeGF \cite{cai_learning_2020} proposed to learn distributions over gradient fields that model shape surfaces.
PointFlow \cite{yang_pointflow_2019} introduced a novel approach using continuous normalizing flows to simultaneously model the distribution of latent variables and the distribution of points for a given shape. SoftFlow \cite{kim_softflow_2020} extended this idea by estimating the conditional distribution of noisy input point clouds perturbed by random noise sampled from various distributions. 

More recently, the advent of DDPMs has led to substantial improvements in 3D point cloud generation. Early diffusion-based methods for point clouds, such as DPM \cite{luo_diffusion_2021}, employed PointNet \cite{qi_pointnet_2017} backbone. Others, including Point-Voxel Diffusion (PVD) \cite{zhou_3d_2021} and LION \cite{zeng_lion_2022}, implemented instead the Point-Voxel Convolutional (PVConv) architecture \cite{liu_point-voxel_2019}. PVD combines a low-resolution voxel-based branch to encode coarse-grained information with a high-resolution point-based branch to capture fine-grained features. LION \cite{zeng_lion_2022} introduced the diffusion in two different latent spaces combining global shape representation with point-structured features. More recently, plain transformer-based diffusion models have gained popularity also for 3D point cloud generation, achieving outperforming results. In particular, DiT-3D \cite{mo_dit-3d_2023} adapted the Diffusion Transformer (DiT) architecture \cite{peebles_scalable_2023} to voxelized point clouds, enabling multi-class training with learnable class embeddings. Similarly, Latent Diffusion Transformer (LDT) \cite{ji_latent_2024} proposed an AE latent compressor to convert raw point clouds into latent tokens, which are then processed by diffusion models. 

As a result, many previous works rely on point encoding techniques such as voxelization, downsampling, or compression, which can degrade the final quality. In contrast, our approach, DiPT, performs diffusion directly on raw point clouds without reducing their resolution, enabling the generation of fine-grained, high-quality samples.


%% file: sec/3_methods.tex
\section{Metrics Rethinking}
\label{sec:metrics}
We identify three key properties for a generative point cloud evaluation metric: (1) invariance to rigid translations of generated samples, (2) consistent behavior across different point distributions, and (3) an inverse monotonic response to noise. The latter property should strictly hold for quality metrics (e.g., MMD), whereas for variability metrics (e.g., COV) we expect invariance at low noise levels and an inverse response only when noise is high enough to alter the underlying shape structure. As shown in \cref{fig:metrics_robustness}, and in more detail in \cref{sec:metrics_analysis} of the Supplementary, one or more of these properties does not always hold for some traditional metrics. For example, MMD-CD and JSD do not exhibit a monotonic inverse response to noise, and none of the traditional metrics are invariant to barycenter shifts.


The proposed enhancements are jointly formalized and validated below, using traditional calculations as a baseline. Analyses are conducted on a set of $4573$ training samples, considered as ideal generations $S_g$, and compared against a reference set $S_r$ of $753$ samples. We progressively introduce random Gaussian noise and/or barycenter shifts proportional to sample diameters (i.e., the maximum inter-point distance), as in \cref{fig:metrics_robustness} (Left). Shapes contain 2,048 points, following the literature \cite{kim_softflow_2020, kim_setvae_2021, zhou_3d_2021, zeng_lion_2022, mo_dit-3d_2023, ji_latent_2024, luo_diffusion_2021, cai_learning_2020}, sampled from the original point clouds either uniformly or randomly in separate trials to simulate uniform and inhomogeneous point distributions.

\begin{figure}
    \centering
    \includegraphics[width=0.9\linewidth]{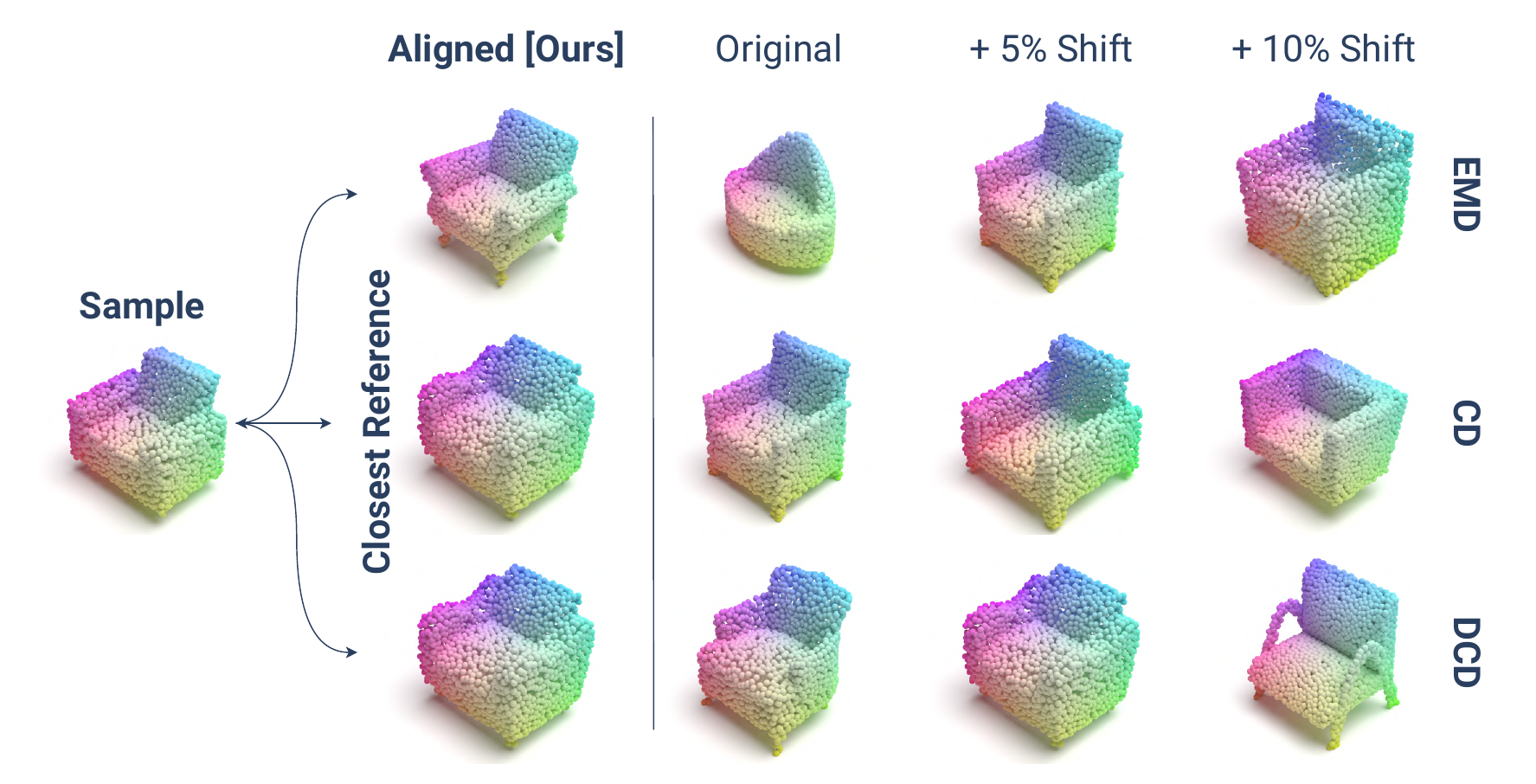}
    \caption{Closest references to a sample under different distance measures with alignment and in response to small shifts.}
    \label{fig:shifts}
\end{figure}

\begin{figure}
    \centering
    \includegraphics[width=.9\linewidth]{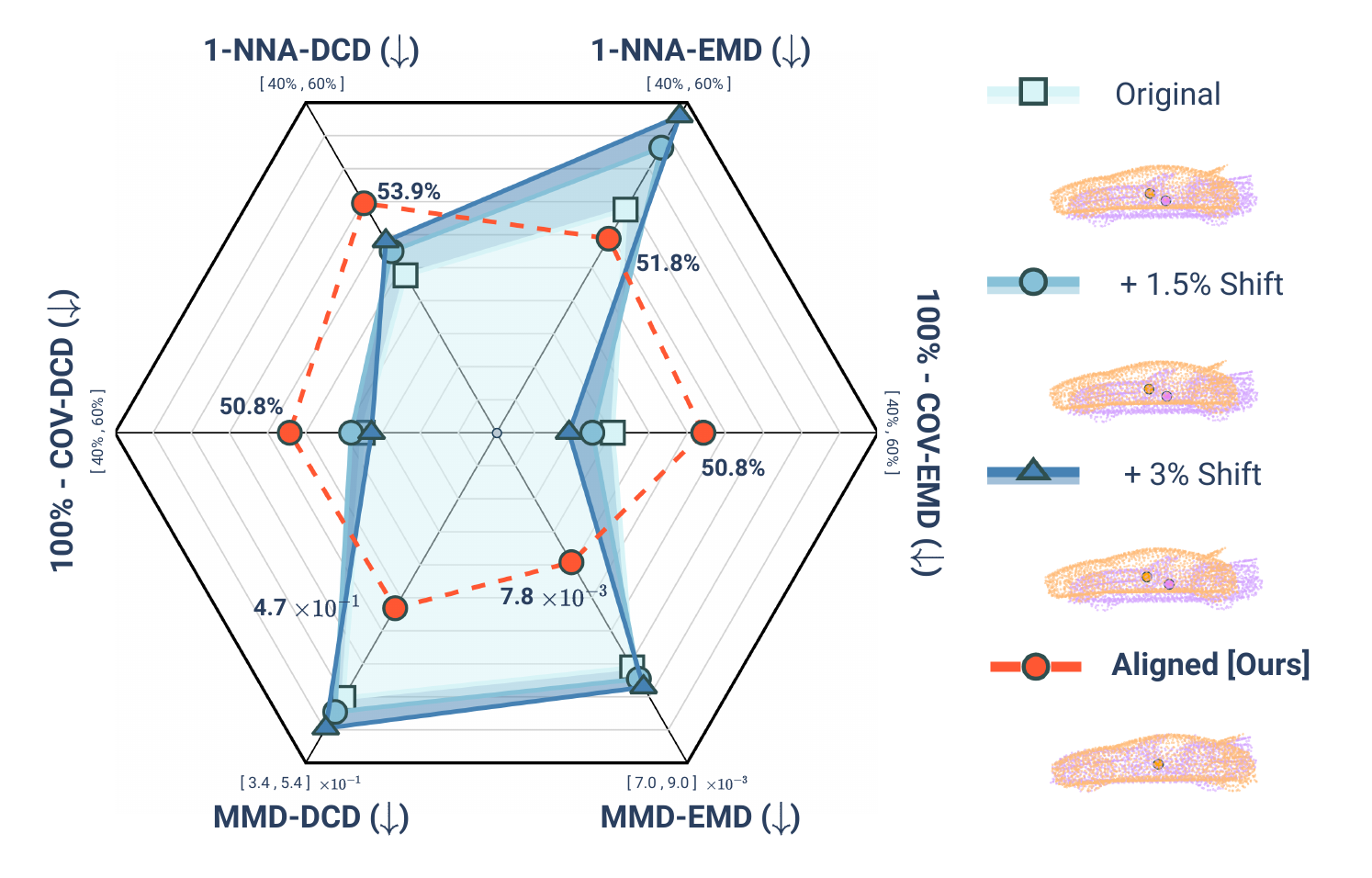}
    \caption{Comparison of 1-NNA, MMD, and COV computed with (red) and without (blue) barycenter alignment. Each metric is evaluated using both DCD and EMD for three levels of shifting.}
    \label{fig:alignementcomparison}
\end{figure}

\textbf{Barycenter Alignment.} Prior works on point cloud generation typically apply global rather than per-sample normalization, using the training set mean and standard deviation \cite{mo_dit-3d_2023, yang_pointflow_2019, zeng_lion_2022, kim_softflow_2020, kim_setvae_2021, zhou_3d_2021}. This ensures the model learns a distribution in normalized space (e.g., varying scales) instead of adapting to each sample specific characteristics. Generated point clouds are eventually de-normalized before evaluating model performances. As a result, barycenters can vary within the same set and between $S_g$ and $S_r$. Furthermore, even with sample-wise normalization, generative models have no theoretical guarantee of producing centered objects, and current evaluation distance measures \cite{achlioptas_learning_2018,yang_pointflow_2019} do not inherently account for such displacements, compromising metric invariance to sample positioning. For example, the same generated point cloud with different small shifts may be matched as closest to different reference samples when alignment is not applied, as in \cref{fig:shifts}, affecting COV and 1-NNA values. To overcome this issue and obtain the desired invariance, we propose a barycenter alignment of point clouds before computing their distances. That is, instead of computing directly a distance measure $D(X, Y)$ between two point clouds, $X=\{ \pmb{x}_i \}_{i=1}^N$ and $Y=\{ \pmb{y}_j \}_{j=1}^M$, we compute 
$D(X-\pmb{x}_b, Y-\pmb{y}_b)$, where $\pmb{x}_b = \frac{1}{N}\sum_{i=1}^N \pmb{x}_i$ and $\pmb{y}_b = \frac{1}{M}\sum_{j=1}^M \pmb{y}_j$. In this way, a generated point cloud with a given structure will always be associated with the same reference regardless of its position in the Euclidean space. A comparison of several metrics computed with and without alignment is shown in \cref{fig:alignementcomparison}. Specifically, the stable metric value achieved using the proposed barycenter alignment is compared to traditional computation, which exhibits undesired variability under small barycenter shifts.

\begin{figure}
    \centering
    \includegraphics[width=0.9\linewidth]{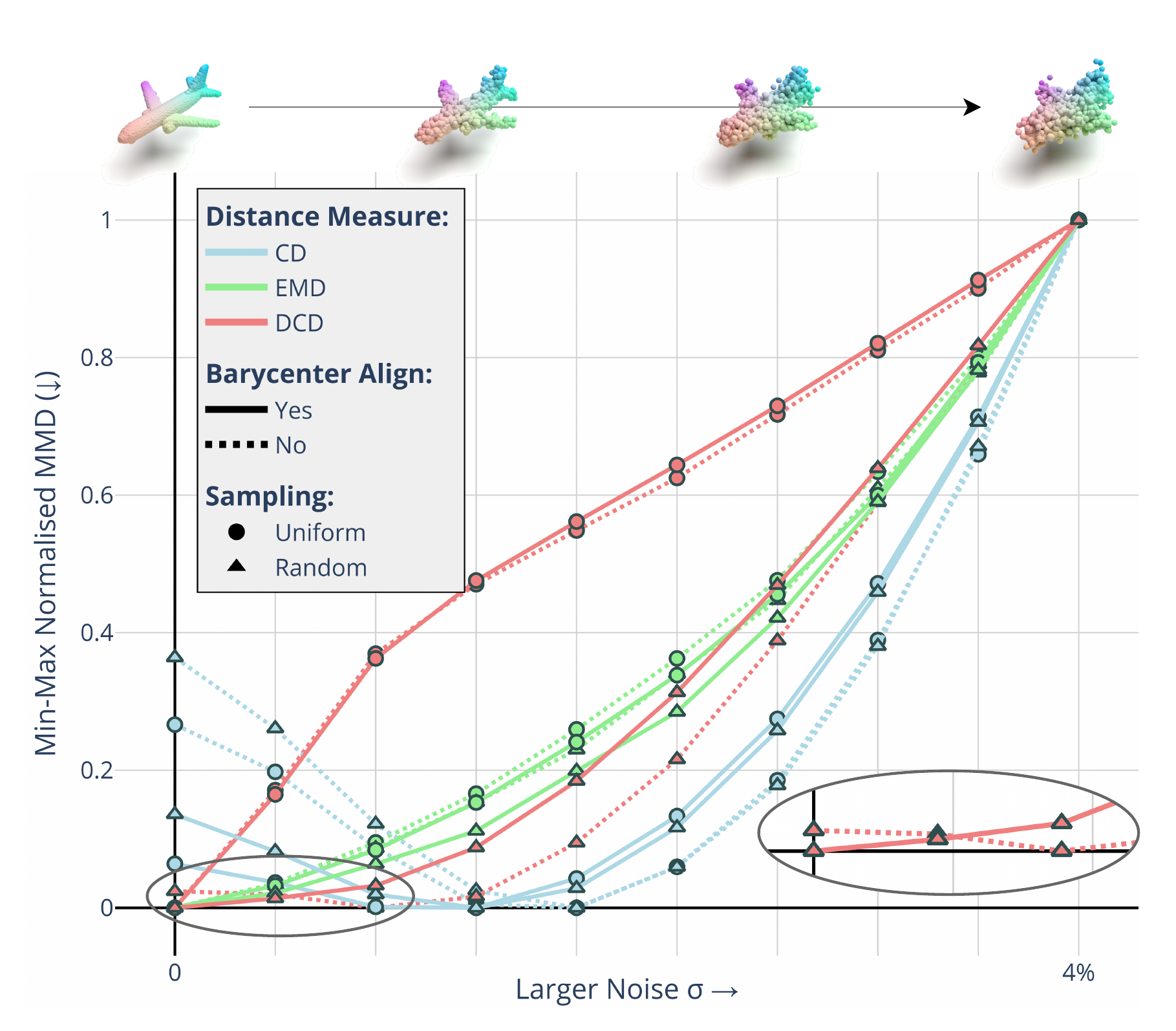}
    \caption{Evolution of the normalized MMD with respect to noise added to the samples of $S_g$, comparing distance measures (CD, EMD, DCD) under different conditions: with or without barycenter alignment and using uniformly or randomly sampled points. }
    \label{fig:mmdvsnoise}
\end{figure}

\textbf{Replacing CD with DCD.} The CD, traditionally used to evaluate generative point cloud models, is well known for its limitations. Among these, it is insensitive to mismatched local density \cite{wu_density-aware_2021}, weakly rotation-aware \cite{liu_self-supervised_2023}, and vulnerable to outliers \cite{lin_hyperbolic_2023}. As a result, CD-based metrics do not always respond inversely to noise. In fact, metrics such as MMD-CD can exhibit improvements when low to mid levels of noise are added to the samples in $S_g$, as shown in \cref{fig:mmdvsnoise}, making them unsuitable as quality indicators. Barycenter alignment mitigates but does not eliminate this issue. To address these limitations, we propose replacing CD in the metrics calculation with the recently introduced DCD \cite{wu_density-aware_2021}, detailed in \cref{eq:DCD} of the Supplementary. DCD is inherited from CD but benefits from a higher sensitivity to distribution quality and has been proven to be a more robust measure of point clouds similarity. These properties make DCD more suitable than CD for evaluating generative models. To validate this intuition, in \cref{fig:mmdvsnoise} we compare the robustness of the MMD metric against the amount of noise added to $S_g$ when computed using different distance measures: CD, EMD, and DCD. Additionally, to cover all scenarios, we compare the metrics computed with and without barycenter alignment for both uniformly and randomly sampled point clouds. In contrast to CD, EMD and DCD demonstrate a monotonically increasing behavior in response to noise. However, MMD-DCD without alignment shows a slight improvement at low noise levels, which disappears once barycenter alignment is applied before distance calculation (see zoom in \cref{fig:mmdvsnoise}). Interestingly, for uniform samples, MMD-DCD increases more rapidly, as perturbations cause stronger density variations that amplify the effect of DCD. This analysis shows that DCD outperforms CD in MMD calculation and underscores the importance of alignment for reliable evaluation of generative models. Intuitively, improving distance calculation with DCD also benefits other metrics, such as 1-NNA and COV. A more detailed comparison between DCD- and CD-based metrics is available in \cref{sec:metrics_analysis} of the Supplementary.

\textbf{Surface Normal Concordance.} 
Several methods have been proposed in the literature for estimating point cloud normals \cite{klasing_comparison_2009}, ranging from the Principal Component Analysis (PCA) of a neighborhood region \cite{hoppe_surface_1992, berkmann_computation_1994}, which is detailed in \cref{sec:normal_estimation} of the Supplementary, to more recent deep learning-based approaches \cite{ben-shabat_nesti-net_2019, guerrero_pcpnet_2018}, as well as other techniques \cite{zhou_fast_2022, wang_weighted_2023, mitra_estimating_2003, pauly_shape_2003}. SNC measures the average similarity of these normals, calculated using any chosen method, between generated samples and their closest references. Specifically, let $M_X \in S_r$ represent the closest reference sample, i.e. the best match, after barycenter alignment, to $X\in S_g$, such that
\begin{equation}
    M_X = \argmin_{Y\in S_r} D(X - \pmb{x}_b, Y - \pmb{y}_b)
\end{equation}
where $D(\cdot, \cdot)$ is any point clouds distance function, e.g. EMD or DCD. Additionally, let $\hat{n}(\cdot)$ denote any method for computing point normals. The SNC is then defined as
\begin{equation}
\label{eq:SNC}
\begin{aligned}
    \text{SNC}(S_r, S_g) &= \frac{1}{|S_g|} \sum_{X \in S_g} \frac{1}{|X|} \sum_{\pmb{x} \in X} \\
    &\quad \bigg|\hat{n}(\pmb{x}) \cdot \hat{n}\big(\argmin_{\pmb{y} \in M_X} ||\pmb{x} - \pmb{y}||_2\big)\bigg|.
\end{aligned}
\end{equation}
Namely, for each point in a generated point cloud, SNC computes the similarity between its normal direction with the normal direction of the closest point from the best-matching shape in the set of references. The proposed metric is highly flexible and can be computed independently of the specific distance measure $D(\cdot,\cdot)$ or normal estimation method $\hat{n}(\cdot)$, as it uses only the absolute value of the cosine to address sign disambiguity, e.g. in PCA-based methods. SNC demonstrates a very strong inverse response to noise, as shown in \cref{fig:metrics_robustness}. This is because small perturbations in point positions cause significant variations in their normals. Thus, SNC is highly sensitive to fine-grained details, making it an ideal complement to traditional metrics for evaluating the quality of generated point clouds. 
Additionally, as discussed in the experiments, normals are independent of global scaling and normalization, enabling fair model comparisons. When computed with a robust method, they are also less sensitive to point distribution than pure Euclidean distances, ensuring consistent behavior.

\begin{figure*}
    \centering
    \includegraphics[width= .75\linewidth]{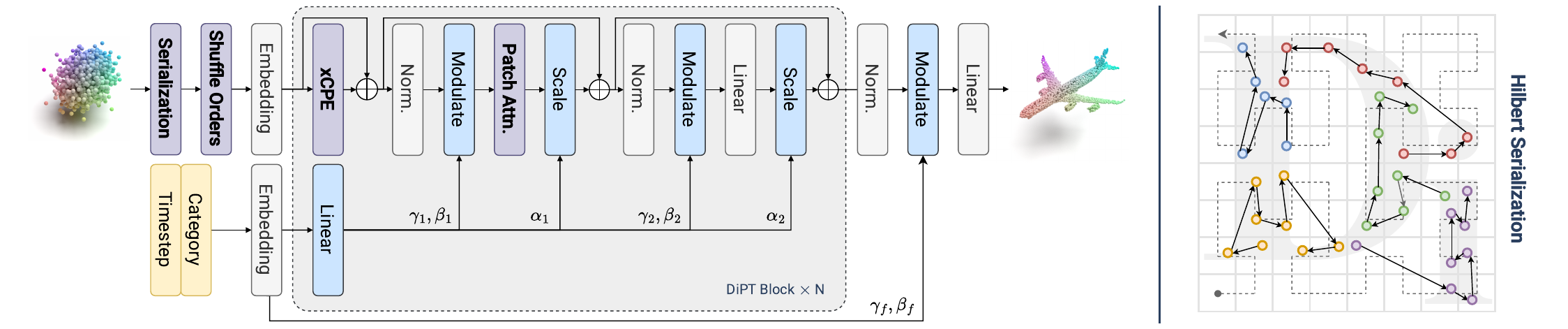}
    \caption{Proposed Diffusion Point Transformer (DiPT) for 3D point cloud generation. (\textbf{Left}) The model serializes the raw input and shuffles the serialization orders before processing it through $N$ DiPT blocks, each performing xCPE, Serialized Patch Attention, and a linear layer. Features are modulated and scaled based on the input condition, composed of the sample category and the diffusion noising timestamp. (\textbf{Right}) Example of Hilbert serialization, where each color represents a patch of maximum size $8$.}
    \label{fig:model}
\end{figure*}

\section{Diffusion Point Transformer}
Inspired by DiT-3D \cite{mo_dit-3d_2023} for its diffusion structure and PTv3 \cite{wu_point_2024} for its backbone architecture, we propose the Diffusion Point Transformer (DiPT) model, in \cref{fig:model}, for 3D point cloud generation. Motivated by some recent advancements \cite{liu_flatformer_2023, wang_octformer_2023, wu_point_2024, wang_serialized_2024}, we transition from the traditional unordered paradigm of point clouds to a serialized structured format. To achieve this, we employ space-filling curves to reorganize point clouds into a one-dimensional sequence by using the Z-order curve \cite{morton_computer_1966}, Hilbert curve \cite{hilbert_dritter_1935}, and their variants Trans-Hilbert and Trans-Z \cite{wu_point_2024}. Importantly, this serialization does not require voxelization nor downsampling. Sparse points are placed into a grid of a given resolution to define the serialized order, as on the right of \cref{fig:model}, allowing the input data to retain its original dimensionality. To enhance generalization capabilities, we incorporate random shuffling of the serialized orders, following the approach of \cite{wu_point_2024, wang_serialized_2024}. This ensures that each DiPT block can learn diverse patterns rather than focusing on a single space-filling curve. Moreover, the serialization enables input points to be grouped into non-overlapping patches, with attention performed independently within each patch, inspired by window attention \cite{liu_swin_2021}. This approach, named Serialized Patch Attention \cite{wu_point_2024}, reduces the computational cost compared to traditional local structure creation methods such as K-Nearest Neighbors (KNN) \cite{zhao_point_2021, wu_point_2022}. Moreover, we replace the absolute sine-cosine embeddings of DiT-3D \cite{mo_dit-3d_2023} or Relative Positional Embeddings (RPE) with enhanced Conditional Positional Encoding (xCPE) \cite{wu_point_2024, wang_octformer_2023}. It consists of a sparse convolution layer with a skip connection before the attention layer of each block, offering more flexibility than traditional positional embeddings for point clouds. As xCPE operates outside the attention mechanism, unlike RPE, it enables optimizations such as flash attention \cite{dao_flashattention-2_2024, dao_flashattention_2022, shah_flashattention-3_2025}, significantly reducing computational time.

Following DiT \cite{peebles_scalable_2023, mo_dit-3d_2023}, we adapt the model for diffusion by incorporating Adaptive Layer Normalization (AdaLN) for feature modulation and scaling based on the input condition. The latter includes time embedding, representing the forward diffusion step, and a learnable class embedding, encoding the category to generate. This design enables multi-class training since in each DiPT block the features scale and shift parameters $\gamma$ and $\beta$ are regressed from the input condition. Additionally, a scaling parameter $\alpha$ is applied after each operation and before residual connections within a block, ensuring condition-dependent feature scaling.

DiPT is designed for scalability, performing point-wise rather than voxel-wise diffusion, and can adapt to different window sizes and model configurations by tuning the patch size and number of blocks. As shown below, it achieves superior point cloud generation quality, producing high-fidelity outputs compared to state-of-the-art methods.

%% file: sec/4_experiments.tex
\section{Experiments}
\begin{figure}
    \centering
    \includegraphics[width=0.80\linewidth]{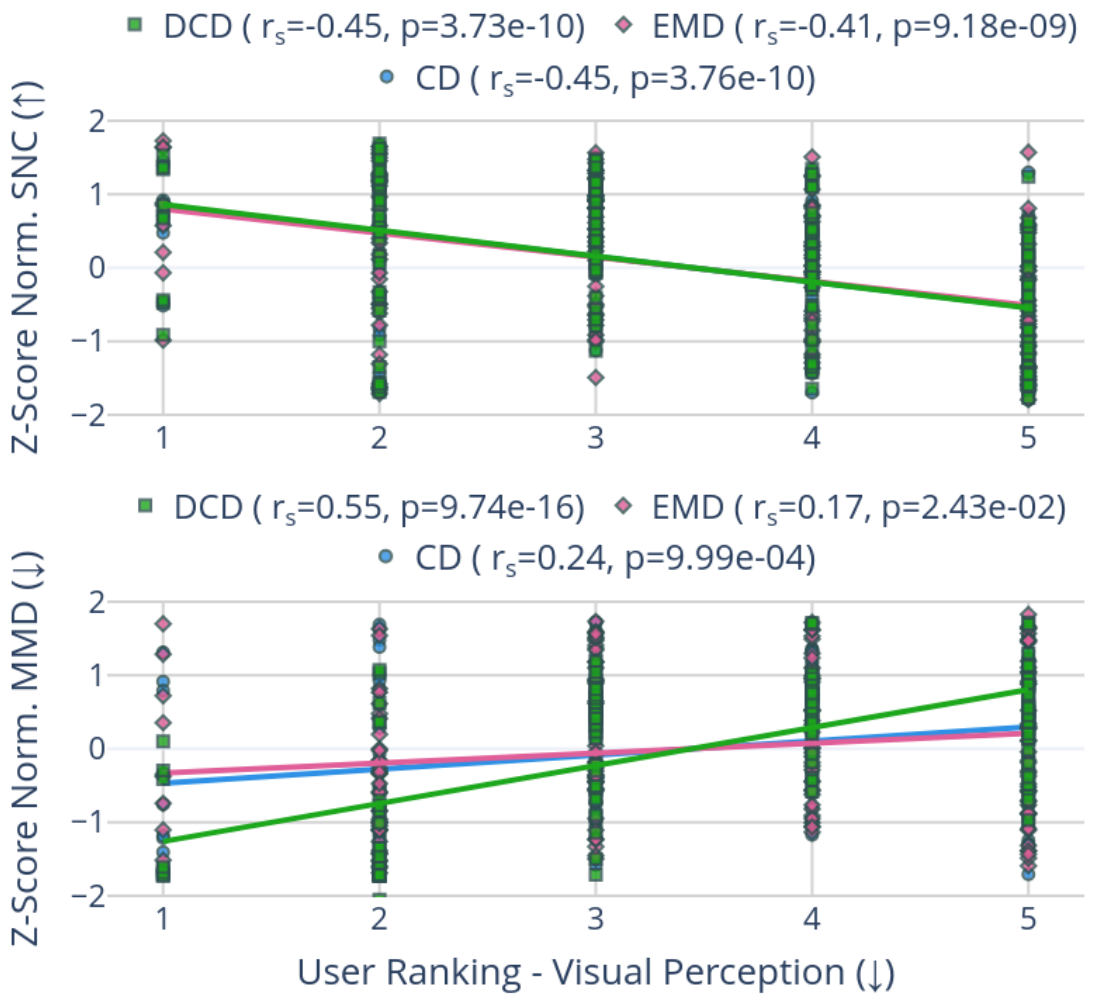}
    \caption{Quality metrics user study. User point clouds perceptual quality rankings are compared to SNC (top) and MMD (bottom). }
    \label{fig:user_study}
\end{figure}
\subsection{Metrics User Study} We conducted a small scale user study to validate the proposed SNC based on human perception of point clouds quality. 15 participants from a mixed audience were asked to sort 5 point clouds, comprising a random reference and its DCD-closest generation from 4 different models, from most to least realistic. Samples were presented in random order via an interactive 3D GUI. Each user ranked samples from the 3 different categories in separate trails, resulting in a total of 45 trials. Correlations between user rankings and quality metrics are shown in \cref{fig:user_study}, with average Spearman scores of $-0.44$ for SNC and $0.32$ for MMD. These results strengthens the proposed SNC by suggesting that it better reflects human perception than MMD. Furthermore, this study confirms the earlier intuition from the MMD analysis in \cref{fig:mmdvsnoise}, with DCD achieving the highest correlation to visual perception among all metrics, and improving CD of $0.31$. In contrast, SNC maintains a similar correlation regardless of the base distance measure, indicating its desired weaker dependence on Euclidean distances.

\subsection{Experiments Settings}
\label{sec:settings}
\textbf{Dataset.} Following previous works \cite{kim_softflow_2020, kim_setvae_2021, zhou_3d_2021, zeng_lion_2022, mo_dit-3d_2023, ji_latent_2024, luo_diffusion_2021, cai_learning_2020}, we used the chair, airplane, and car categories from ShapeNet \cite{chang_shapenet_2015} to train the DiPT model. For each training shape, we sampled 2048 points using Furthest Point Sampling (FPS). We adopted the same dataset splits and pre-processing steps introduced in PointFlow \cite{yang_pointflow_2019}, which are widely adopted in the community \cite{kim_softflow_2020, kim_setvae_2021, zhou_3d_2021, zeng_lion_2022, mo_dit-3d_2023}, including global sample normalization. Additional DiPT experiments on 10 mixed ShapeNet categories, as well as ablations on model size and components (e.g., positional embeddings), are provided in \cref{sec:additional_experiments} of the Supplementary.

\begin{table*}[t]
\caption{Comparison of metrics across different models for 3D point cloud generation. All models are evaluated using the same uniformly sampled reference set and their public generated samples or weights. MMD is omitted for models trained with different input normalization, as it does not provide a fair comparison. The best scores are highlighted in bold. MMD-DCD is scaled by $10$, and MMD-EMD by $10^3$.}
\label{tab:model_comparison}
\footnotesize
\begin{tabularx}{\textwidth}{ll|YY|YY|YY|YY}
\toprule 
 &  \multirow{3}{*}{\vspace{-3.5mm}Model} & \multicolumn{4}{c|}{\textbf{Variability}} & \multicolumn{4}{c}{\textbf{Quality}} \\
 \cmidrule{3-10}
 &  & \multicolumn{2}{c|}{1-NNA (\%,$\downarrow$)} & \multicolumn{2}{c|}{COV (\%,$\uparrow$)} & \multicolumn{2}{c|}{MMD ($\downarrow$)} & \multicolumn{2}{c}{SNC (\%, $\uparrow$)}\\ 
\cmidrule{3-4} \cmidrule{5-6} \cmidrule{7-8} \cmidrule{9-10}
     & & DCD & EMD & DCD & EMD & DCD & EMD & DCD & EMD \\ 
    \midrule

\multirow{9}{*}{\rotatebox[origin=c]{90}{Chair}} & PointFlow \cite{yang_pointflow_2019} & 60.72 & 60.18 & 43.64 & \textbf{52.07} & 6.49 & 8.91 & 70.93 & 69.42 \\ 
& SoftFlow \cite{kim_softflow_2020} & 61.64 & 67.76 & 37.67 & 43.34 & 6.47 & 9.07 & 73.11 & 71.48 \\
& ShapeGF \cite{cai_learning_2020} & 55.28 & 64.47 & \textbf{49.16} & 45.48 & - & - & 73.99 & 73.08 \\
& SetVAE \cite{kim_setvae_2021} & 62.33 & 66.54 & 43.19 & 41.04 & 6.44 & 8.73 & \textbf{77.41} & 74.61 \\
& DPM \cite{luo_diffusion_2021} & 70.21 & 91.65 & 40.12 & 33.84 & - & - & 70.14 & 68.21 \\
& PVD \cite{zhou_3d_2021} & 52.60 & \textbf{54.13} & 45.33 & 48.24 & 6.46 & \textbf{8.46} & 75.94 & 73.72 \\
& LION \cite{zeng_lion_2022} & \textbf{51.61} & 54.98 & 44.72 & 49.46 & 6.44 & 8.54 & 75.47 & 73.19 \\
& DiT-3D \cite{mo_dit-3d_2023} & 99.00 & 91.35 & 17.00 & 19.14 & 6.68 & 9.85 & 76.12 & 73.52 \\ 
& \textbf{DiPT [Ours]} & 68.68 & 64.47 & 41.81 & 43.95 & \textbf{6.08} & 8.47 & 77.29 & \textbf{75.10} \\
\midrule
\multirow{9}{*}{\rotatebox[origin=c]{90}{Airplane}} & PointFlow \cite{yang_pointflow_2019} & 66.67 & 86.30 & 40.99 & 38.27 & 4.30 & 2.35 & 83.25 & 81.12 \\
& SoftFlow \cite{kim_softflow_2020} & 66.79 & 90.37 & 40.00 & 38.52 & 4.26 & 2.40 & 84.05 & 81.98 \\
& ShapeGF \cite{cai_learning_2020} & 64.94 & 92.10 & 47.41 & 30.86 & - & - & 83.27 & 81.32 \\
& SetVAE \cite{kim_setvae_2021} & 64.69 & 88.52 & 38.52 & 36.79 & 4.26 & 2.24 & 87.39 & 85.51 \\
& DPM \cite{luo_diffusion_2021} & 68.40 & 92.96 & 40.99 & 28.15 & - & - & 82.86 & 81.20 \\
& PVD \cite{zhou_3d_2021} & \textbf{60.62} & 82.35 & 43.46 & 40.00 & 4.36 & 2.17 & 84.42 & 82.60 \\
& LION \cite{zeng_lion_2022} & 65.68 & 84.94 & 44.44 & 39.01 & 4.24 & 2.30 & 83.01 & 80.90 \\
& LDT \cite{ji_latent_2024} & 90.25 & 90.86 & 44.20 & 34.32 & - & - & 86.35 & 83.95 \\ 
&\textbf{DiPT [Ours]} & 63.70 & \textbf{74.32} & \textbf{44.20} & \textbf{46.42} & \textbf{3.29} & \textbf{1.65} & \textbf{87.50} & \textbf{86.00} \\
\midrule
\multirow{8}{*}{\rotatebox[origin=c]{90}{Car}} & PointFlow \cite{yang_pointflow_2019} & 50.85 & 61.97 & 43.02 & 49.00 & 5.41 & 3.69 & 76.84 & 74.67 \\ 
& SoftFlow \cite{kim_softflow_2020} & \textbf{50.57} & 67.38 & 37.89 & 45.01 & 5.39 & 3.75 & 78.31 & 75.84 \\
& ShapeGF \cite{cai_learning_2020} & 52.71 & 68.23 & 46.72 & 45.01 & - & - & 77.62 & 75.69 \\
& SetVAE \cite{kim_setvae_2021} & 53.42 & 72.65 & 36.47 & 49.29 & 5.38 & 3.55 & 82.54 & 79.82 \\
& PVD \cite{zhou_3d_2021} & 50.71 & 64.25 & 42.74 & 51.28 & 5.55 & 4.54 & 78.99 & 76.45 \\
& LION \cite{zeng_lion_2022} & 50.85 & 64.39 & 41.88 & \textbf{53.28} & 5.48 & 3.70 & 78.01 & 75.73 \\
& LDT \cite{ji_latent_2024} & 75.93 & 73.08 & \textbf{47.86} & 50.43 & - & - & 82.26 & 78.89 \\ 
&\textbf{DiPT [Ours]} & 61.11 & \textbf{60.26} & 36.47 & 44.44 & \textbf{4.65} & \textbf{3.28} & \textbf{82.69} & \textbf{80.64} \\
\bottomrule
\end{tabularx}
\end{table*}

\textbf{Implementation Details.} For comparison with other methods, we trained the proposed DiPT model following the Small (S) ViT and DiT architecture \cite{mo_dit-3d_2023, peebles_scalable_2023, dosovitskiy_image_2021}. Namely, we used $12$ blocks with feature size $384$ and $6$ attention heads. Inspired by Swin-Transformer \cite{liu_swin_2021}, we alternate small and large patch sizes for the serialized attention, repeating the pattern $256$ - $512$ - $1024$ - $1024$ and aiming to capture both local and global information relevant for generation variability and quality, respectively. The models were trained on $32$ NVIDIA H100 GPUs for $10000$ epochs using the AdamW optimizer \cite{loshchilov_decoupled_2019} and one-cycle learning rate policy \cite{smith_super-convergence_2019} with a maximum learning rate of $2\mathrm{e}{-4}$. Finally, we used a DDPM scheduler with $1000$ noising steps with linearly increasing forward process variances from $1\mathrm{e}{-4}$ to $0.02$, as in \cite{ho_denoising_2020}. SNC was calculated using PCA-based normals \cite{hoppe_surface_1992, berkmann_computation_1994} extracted from neighborhoods of $20$ points. We found this value to be a good trade-off between local and global normal information for the ShapeNet samples (see \cref{fig:SNC_param} of the Supplementary). We chose this method for its simplicity and flexibility in handling varying distributions, as it focuses on local geometric structures.

\subsection{Experiments Results} 
\label{sec:results}
\begin{figure*}
    \centering
    \includegraphics[width=.93\linewidth]{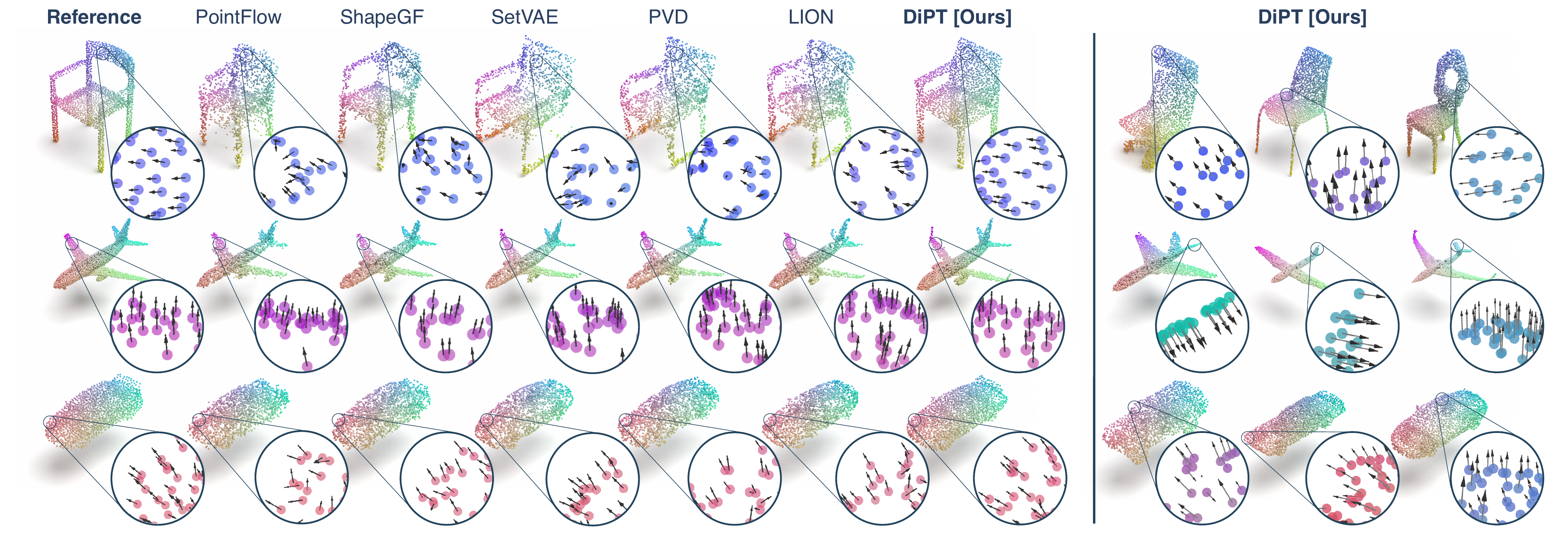}
    \caption{(\textbf{Left}) Qualitative comparison of the closest generated 3D point clouds to the reference based on DCD, across different models for the chair, airplane, and car categories. Additionally, point normals in zoomed regions are shown for samples smoothness comparison. (\textbf{Right}) Additional samples generated by DiPT.}
    \label{fig:comparison}
\end{figure*}

We present, in \cref{tab:model_comparison}, a quantitative comparison of generative models using the proposed enhanced evaluation metrics. Note that JSD is excluded from the analysis, as it remains the only metric that lacks robustness and stability, even after the refinements (see \cref{fig:JSD} of the Supplementary). Our DiPT model demonstrates its superiority over the others, achieving the best performance on qualitative metrics MMD and SNC across all categories. Compared to DiT-3D with the same Small (S) model size \cite{mo_dit-3d_2023}, our model demonstrates significantly better generalization (greater variability) while simultaneously producing higher-quality samples. Furthermore, the introduced SNC metric complements MMD by providing a deeper understanding of the quality of the generated samples. When MMD cannot be compared fairly due to different normalization, and consequently different generated point cloud sizes, SNC can be used as the only reliable quality indicator, as it is not affected by scale. Additionally, when MMD values are very close or discordant when computed with different distance measures, SNC helps in better interpreting the results. For example, in airplane generation, SetVAE \cite{kim_setvae_2021} and PVD \cite{zhou_3d_2021} exhibit discordant MMD-DCD and MMD-EMD values, with one model outperforming the other on only one measure. The SNC metric, however, reveals that SetVAE produces higher-quality samples, as its value is consistently higher for both DCD and EMD. In fact, as shown in the graphical comparison in \cref{fig:comparison}, the airplane sample generated by SetVAE seems less noisy than the one generated by PVD, in concordance with MMD-DCD and SNCs. The figure also illustrates the superiority of our DiPT model, which generates high-fidelity samples with sharper contour definitions and smoother normals compared to those of other models. In terms of variability, measured by the 1-NNA and COV metrics, the proposed DiPT outperforms the other methods in the airplane category. However, for the other categories, no single method clearly outperforms the others. This is expected, as variability metrics depend solely on the diversity within each category and can fluctuate in the presence of noise. Nevertheless, DiPT outperforms in 4 out of 12 scores (3 on airplane, 1 on car). LION leads in 2, while others top only 1. As the overall best in quality, DiPT thus also offers the best variety-quality tradeoff among models.

\subsection{Discussion} In this work, since the ShapeNet data share the same orientation, we introduced only barycenter alignment for simplicity. Nevertheless, as all traditional metrics, SNC is also sensitive to rotation and geometry, therefore a rigid registration method like ICP or CPD \cite{myronenko_non-rigid_2006} might be required in more general scenarios, before computing point cloud distances, to handle rotation mismatches, e.g. on DiPT-S, SNC improves in mean $0.06\%$ with ICP but runs $3.35\times$ slower. Moreover, \cref{tab:model_comparison_random} in the Supplementary presents the same comparison as in \cref{tab:model_comparison}, but with inhomogeneous references. The results show that SNCs, along with MMD-DCD, are the most consistent metrics for preserving relative model rankings across different reference distributions, achieving the highest rank correlations. This supports SNC's reliability despite geometry mismatches, provided a consistent reference set is used. Furthermore, the proposed SNC is designed for single objects where the evaluation of surface smoothness is of interest. Consequently, it may struggle with irregular objects, such as trees, and may require tuning of the normal estimation techniques, e.g. by changing the neighbors region size for PCA or dynamically adapting them based on object complexity. SNC is analyzed in details under mismatched point densities and different normal estimation settings in \cref{sec:normal_estimation} of the Supplementary. Additionally, for generating scenes, such as in LiDAR sequences \cite{behley_semantickitti_2019}, SNC can still be used, along with other metrics \cite{triess_realism_2022}, by decomposing the scene into smaller objects, such as cars, pedestrians, and buildings, and evaluating the surface quality of each compared to a set of references.

%% file: sec/5_conclusions.tex
\section{Conclusions}
We introduced new guidelines to ensure a more reliable assessment of 3D point cloud generative models by enhancing the fidelity of evaluation metrics in reflecting the true quality of generated samples, making them robust to shifts and more sensitive to defects such as noise. Additionally, we introduced the SNC metric to evaluate the surface quality of generated samples by comparing their estimated point normals with those of the references. We believe that the proposed SNC can help assess, and consequently improve, the quality of synthesized shapes by complementing MMD in cases where it struggles and particularly when surface regularity is of primary interest. When normals are less relevant, our work encourages future metrics to target other meaningful properties as needed. Furthermore, the proposed DiPT model combines innovations from point cloud processing and diffusion models, outperforming previous methods in generative quality, as shown on the ShapeNet dataset. Our framework strengthens evaluation methods and opens avenues for further research in 3D generation. Advancing these techniques could lead to more accurate, realistic, and consistent 3D models. A promising direction for future work is to adapt the proposed model and metrics to other fields, such as LiDAR scans or domain-specific datasets, while also dynamically adjusting metrics like SNC based on shape complexity, irregularities, and requirements, leading to more generalizable assessments of 3D generative models.

%% file: sec/6_ack.tex
\section{Acknowledgements}
This project has received funding from the European Union’s Horizon 2020 research and innovation programme under the Marie Skłodowska-Curie grant agreement No 945304-Cofund AI4theSciences hosted by PSL University.
This work was granted access to the HPC/AI resources of IDRIS under the allocation 2022-AD011013902 made by GENCI.

%% file: sec/X_suppl.tex
\maketitlesupplementary
\setcounter{page}{1}

\section{Assessment of Point Cloud Generation}
\label{sec:metrics_details}
\begin{figure*}
    \centering    
    \includegraphics[width=0.8\linewidth]{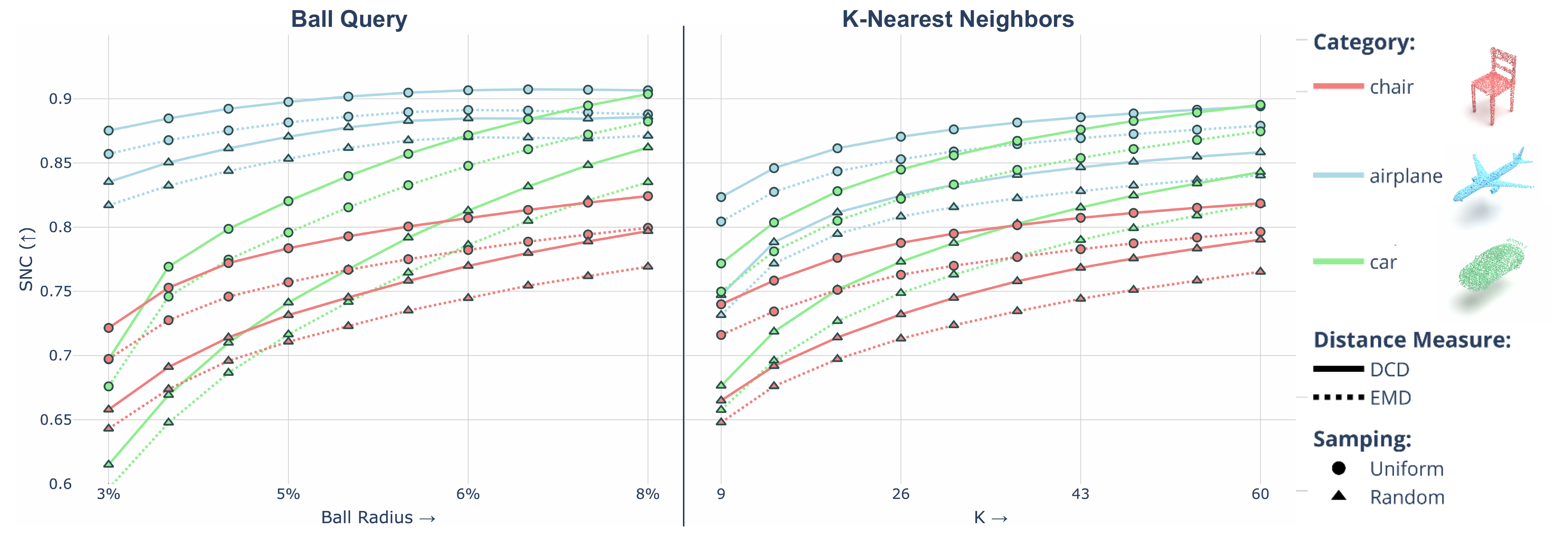}
    \caption{SNC metric using two different neighbor selection methods for normal estimation: Ball Query and K-Nearest Neighbors. For each method, the metrics evolution is shown with respect to their selection parameter, i.e., ball radius for Ball Query and number of neighbors $K$ for K-Nearest Neighbors. Moreover, SNC is evaluated for three different classes (chair, car, and airplane) under various conditions: using either DCD or EMD distance measures and employing uniform or random sampling of points.}
    \label{fig:SNC_param}
\end{figure*}
\begin{figure*}
    \centering    
    \includegraphics[width=\linewidth]{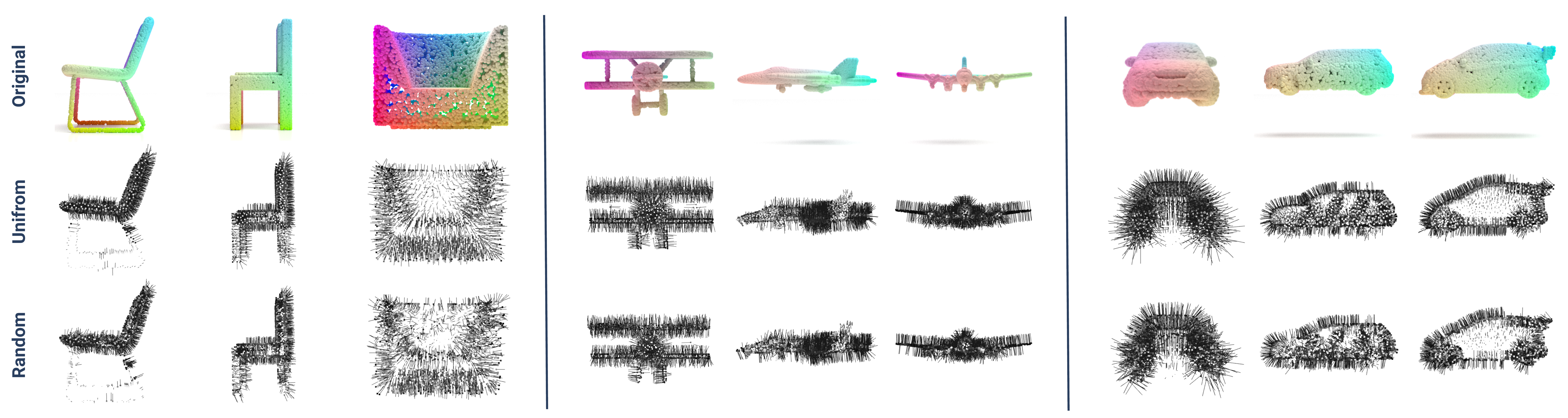}
    \caption{Graphical comparison of normals estimated using 3D plane fitting through PCA method and KNN selection with $K=20$ when selecting $2048$ points from the original point clouds using uniform and random sampling.}
    \label{fig:normals}
\end{figure*}
\textbf{Distance Measures.} Following previous works \cite{yang_pointflow_2019, mo_dit-3d_2023, zhou_3d_2021, achlioptas_learning_2018}, Chamfer Distance (CD) and Earth Mover Distance (EMD) are often used to measure similarity between point clouds. Let $X=\{ x_i \}_{i=1}^N$ and $Y=\{ y_j \}_{j=1}^M$ be two arbitrary point clouds, we can formally define:
\begin{equation}
    \label{eq:CD}
    \text{CD}(X,Y) = \sum_{x \in X} \min_{y \in Y} || x -y ||_2^2 + \sum_{y \in Y} \min_{x \in X} || x -y ||_2^2 
\end{equation}
\begin{equation}
    \label{eq:EMD}
    \text{EMD}(X,Y) = \min_{\phi: X \rightarrow Y} \sum_{x \in X}||x- \phi(x)||_2
\end{equation}
where $\phi$ is a bijection between $X$ and $Y$ when $|X| = |Y|$. To tackle well-known limitations of CD, such as insensitivity to mismatched local density \cite{wu_density-aware_2021}, weakly rotation-awareness \cite{liu_self-supervised_2023}, and vulnerability to outliers \cite{lin_hyperbolic_2023}, \textit{Wu et al.} introduced Density-Aware Chamfer Distance (DCD) \cite{wu_density-aware_2021}. DCD is inherited from CD but benefits from a higher sensitivity to distribution quality and has been proven to be a more robust measure of point cloud similarity. DCD is defined as

\begin{equation}
\label{eq:DCD}
\begin{aligned}
    \text{DCD}(X, Y) &= \frac{1}{2} \Bigg( 
    \frac{1}{|X|} \sum_{x \in X} \left( 1 - \frac{1}{n_{\hat{y}}} e^{-\alpha \|x - \hat{y}\|_2} \right) \\
    &\quad + \frac{1}{|Y|} \sum_{y \in Y} \left( 1 - \frac{1}{n_{\hat{x}}} e^{-\alpha \|y - \hat{x}\|_2} \right) 
    \Bigg)
\end{aligned}
\end{equation}
where $\hat{y} = \min_{y \in Y} \|x - y\|_2$, $\hat{x} = \min_{x \in X} \|y - x\|_2$, and $\alpha$ denotes a temperature scalar. Additionally, $n_{\hat{x}}$ and $n_{\hat{y}}$ are the number of points that query $\hat{x}$ and $\hat{y}$, i.e. the number of points for which the closest points are $\hat{x}$ and $\hat{y}$, respectively. 
 
\textbf{Evaluation Metrics.} When it comes to structured data, such as graphs and 3D point clouds, the focus of generative evaluation metrics is to compare the structural properties of generated and real data. Let $S_g$ be the set of generated point clouds and $S_r$ be the set of reference point clouds with $|S_r| = |S_g|$. Instead of directly considering distance measures between samples as metrics to evaluate generative models, \textit{Achlioptas et al.} \cite{achlioptas_learning_2018} introduced three different metrics:
\begin{itemize}
    \item \textbf{Coverage (COV)} measures the fraction of point clouds in the reference set that can be associated to at least one point cloud in the generated set. For that purpose, each point cloud in $S_g$ is matched to the closest in $S_r$ according to a distance metric $D(\cdot, \cdot)$:
    \begin{equation}
        \text{COV}(S_r, S_g) = \frac{|\{\arg\min_{Y\in S_r} D(X,Y) | X \in S_g|\}}{|S_r|}.
    \end{equation} 
    In other words, the coverage measures how different the generated samples are according to the variability of the reference set. Nevertheless, it is only a measure of diversity of the generated point clouds, but it does not capture their quality.
    \item \textbf{Minimum Matching Distance (MMD)} is therefore proposed as a metric that measures quality. For each point cloud in $S_r$, the distance from its nearest neighbor in $S_g$ is calculated and averaged:
    \begin{equation}
        \text{MMD}(S_r, S_g) = \frac{1}{|S_r|} \sum_{Y \in S_r} \min_{X \in S_g} D(X, Y).
    \end{equation}
    However, only a few good generated samples are needed to obtain low MMD values, overshadowing possible low-quality point clouds. In fact, the same high-quality generated sample can be the best match of multiple elements in $S_r$ and bad samples may never participate in the metric calculation.
    \item \textbf{Jensen-Shannon Divergence (JSD)} is computed between the marginal point distributions
    \begin{equation}
    \label{eq:JSD}
        \text{JSD}(S_g, S_r) = \frac{1}{2}D_{\text{KL}}(P_r||M) + \frac{1}{2}D_{\text{KL}}(P_g||M)
    \end{equation}
    where $M=\frac{1}{2}(P_r + P_g)$ and $P_r$ and $P_g$ are marginal distributions of points in the $S_g$ and $S_r$ obtained by assigning each point to a voxel of the voxelized input space using a given voxel size $V$, and $D_{\text{KL}}(\cdot||\cdot)$ is the Kullback–Leibler (KL)-divergence \cite{kullback_information_1951} between the two distributions. This metric is very basic since it works with marginals and not distributions of individual samples and therefore also has several limitations.
\end{itemize}
To overcome the drawbacks and limitations of the previous metrics, \textit{Yang et al.} \cite{yang_pointflow_2019} introduced another metric better suited for the evaluation of point clouds generative models: the \textbf{1-nearest neighbor accuracy (1-NNA)}. It was originally proposed for two-sample tests \cite{lopez-paz_revisiting_2017}, but it was also adapted to evaluate the performance of Generative Adversarial Networks (GANs) \cite{xu_empirical_2018}. For point clouds evaluation, 1-NNA is defined as
\begin{equation}
    \begin{split}
        &\text{1-NNA}(S_g, S_r) = \\ 
        &\frac{\sum_{X \in S_g}\mathbb{I}[N_X \in S_g] + \sum_{Y \in S_r}\mathbb{I}[N_Y \in S_r]}{|S_g| + |S_r|}
    \end{split}
\end{equation}
where $N_X$ is the nearest neighbor of $X$ in $S_{-X} = S_r \cup S_g - {X}$ computed using any $D(\cdot, \cdot)$ and $\mathbb{I}$ is the indicator function. In other words, each sample of $S_g \cup S_r$ is classified as belonging to $S_g$ or $S_r$ based on its nearest neighbor. Therefore, if $S_g$ and $S_r$ are sampled from the same distribution, then 1-NNA is likely to converge to $50\%$ since nearest neighbors of samples should belong to either sets with equal probability. Therefore, 1-NNA directly accounts for shape distributions (unlike JSD, which considers marginals) and should reflect both the diversity and fidelity of generated samples  simultaneously.

The four introduced metrics, COV, MMD, JSD and 1-NNA, have been consistently used together in previous works \cite{kim_softflow_2020, kim_setvae_2021, zhou_3d_2021, zeng_lion_2022, mo_dit-3d_2023, ji_latent_2024, luo_diffusion_2021, cai_learning_2020} to assess the performance of point cloud generative models, aiming to capture both variability and quality aspects.

\section{Point Cloud Normal Estimation}
\label{sec:normal_estimation}
Though many different normal estimation methods exist \cite{klasing_comparison_2009}, the simplest approach is based on first-order 3D plane fitting within a neighborhood of points, as proposed by \textit{Hoppe et al.} \cite{hoppe_surface_1992} and \textit{Berkmann et al.} \cite{berkmann_computation_1994}. Therefore, determining the normal at a point of a point cloud can be approximated by estimating the normal of a plane tangent to the surface, which reduces to a least-squares plane fitting problem. Consequently, the solution for estimating the surface normal at a point $\pmb{x} \in X$ is equivalent to performing Principal Component Analysis (PCA) on the covariance matrix constructed from a set of its neighbors, $\mathcal{N}(\pmb{x})$, and analyzing its eigenvectors and eigenvalues. The most commonly used methods to define the set of neighbors $\mathcal{N}(\pmb{x})$ are:
\begin{itemize}  
    \item \textbf{K-Nearest Neighbors (KNN)}: Select the $K$ nearest points to $ \pmb{x} $.  
    \item \textbf{Ball Query}: Select the points within a sphere of radius $r$ centered at $ \pmb{x} $.  
\end{itemize}  
KNN ensures a fixed number of neighbors, which is useful for consistency but may include distant points in sparse areas. Ball Query adapts to local density but can result in a varying number of neighbors, which may be less stable. Therefore, the choice and tuning of the neighbor selection method depend on the application and should be adjusted based on the characteristics of the analyzed point clouds.  

For each point $\pmb{x} \in X$ and its set of neighbors $ \mathcal{N}(\pmb{x}) $, the covariance matrix is defined as 

\begin{equation}
    \mathcal{C}_{\pmb{x}} = \frac{1}{|\mathcal{N}(\pmb{x})|} \sum_{\pmb{p} \in \mathcal{N}(\pmb{x})} (\pmb{p} - \overline{\pmb{p}}) \cdot (\pmb{p} - \overline{\pmb{p}})^T
\end{equation}
where $\overline{\pmb{p}}$ represents the centroid of the points in $\mathcal{N}(\pmb{x})$. $\mathcal{C}_{\pmb{x}}$ is symmetric and positive semi-definite; therefore, its eigenvalues are real and non-negative. The eigenvectors  
\begin{equation}  
    \mathcal{C}_{\pmb{x}} \pmb{\phi}_j = \lambda_j \pmb{\phi}_j  
\end{equation}  
for $j \in \{1,2,3\}$ form an orthogonal frame. If the eigenvalues satisfy $0 \leq \lambda_0 \leq \lambda_1 \leq \lambda_2$, then the eigenvector $\pmb{\phi}_0$, corresponding to the smallest eigenvalue $\lambda_0$, provides an approximation of the desired normal at the point $\pmb{x}$.

Nevertheless, the orientation of the normal computed through PCA is ambiguous and may not be consistent across the entire point cloud $X$. This issue can be easily addressed by orienting all normals consistently towards the viewpoint, provided it is known. A key advantage of the proposed SNC metric in \cref{eq:SNC} is that this step is unnecessary, as the metric relies solely on the angle between directions, making their orientations irrelevant for its computation.

In \cref{fig:SNC_param}, the evolution of the SNC metric is shown for the chair, airplane, and car categories in different scenarios, based on the neighbor selection method and its selection parameter. For the ball query method, we vary the ball radius between $3\%$ and $8\%$ of the samples diameter, which correspond on average to approximately $9$ to $60$ neighbors. Overall, SNC values increase as the region for 3D plane fitting and normal calculation expands. This is expected because, with fewer points, the normals capture more local information, making matching more challenging. On the other hand, when a relatively large number of points is used, the normals become smoother and more uniform, facilitating matching. Therefore, depending on the desired precision and the complexity of the generated shape, the neighbor querying parameter, $K$ or $r$, can be tuned accordingly. Moreover, SNC exhibits consistent behavior when computed using different distance measures, such as DCD and EMD, ensuring a more robust evaluation of generated sample quality. Finally, when point clouds are in-homogeneous, i.e. when using random sampling, the SNC metric is generally lower than with uniform point clouds but still maintains the same trend. This drop occurs because, in in-homogeneous point clouds, the density of points varies across the surface, leading to less reliable normal estimations in sparse regions. As a result, normals become more irregular, making their correct matching more challenging compared to uniformly sampled point clouds, where normal estimation is more stable and precise. \cref{fig:normals} shows a graphical comparison of point normals estimated on the same point cloud, sampled both uniformly and randomly from the original data. Normals estimated on in-homogeneous point clouds are slightly noisier compared to those on uniformly sampled ones. Importantly, in both cases, they remain consistent with the analyzed surface and can be reliably used to calculate the SNC metric.

\section{Detailed Metrics Analysis}
\label{sec:metrics_analysis}
In the following, we analyze the response of each metric, including traditional methods (JSD, MMD, COV, and 1-NNA) and our proposed SNC, to increasing levels of noise and sample shifts. This evaluation is conducted across four distinct scenarios, defined by:
\begin{itemize}
\item \textbf{Sampling Strategy}: Each point cloud consists of 2048 points, selected from the original samples either through uniform sampling (resulting in a homogeneous point distribution) or random sampling (leading to an in-homogeneous point distribution).
\item \textbf{Alignment Approach}: Metrics are computed either using the traditional approach or with the proposed barycenter alignment, which aims to improve robustness against shifts.
\end{itemize}
Additionally, when metrics are based on pair-wise point cloud distances, we analyze their response using the three different distance measures introduced in \cref{eq:CD}, \cref{eq:EMD}, and \cref{eq:DCD}: CD, EMD, and DCD.

\begin{figure}
    \centering    
    \includegraphics[width=\linewidth]{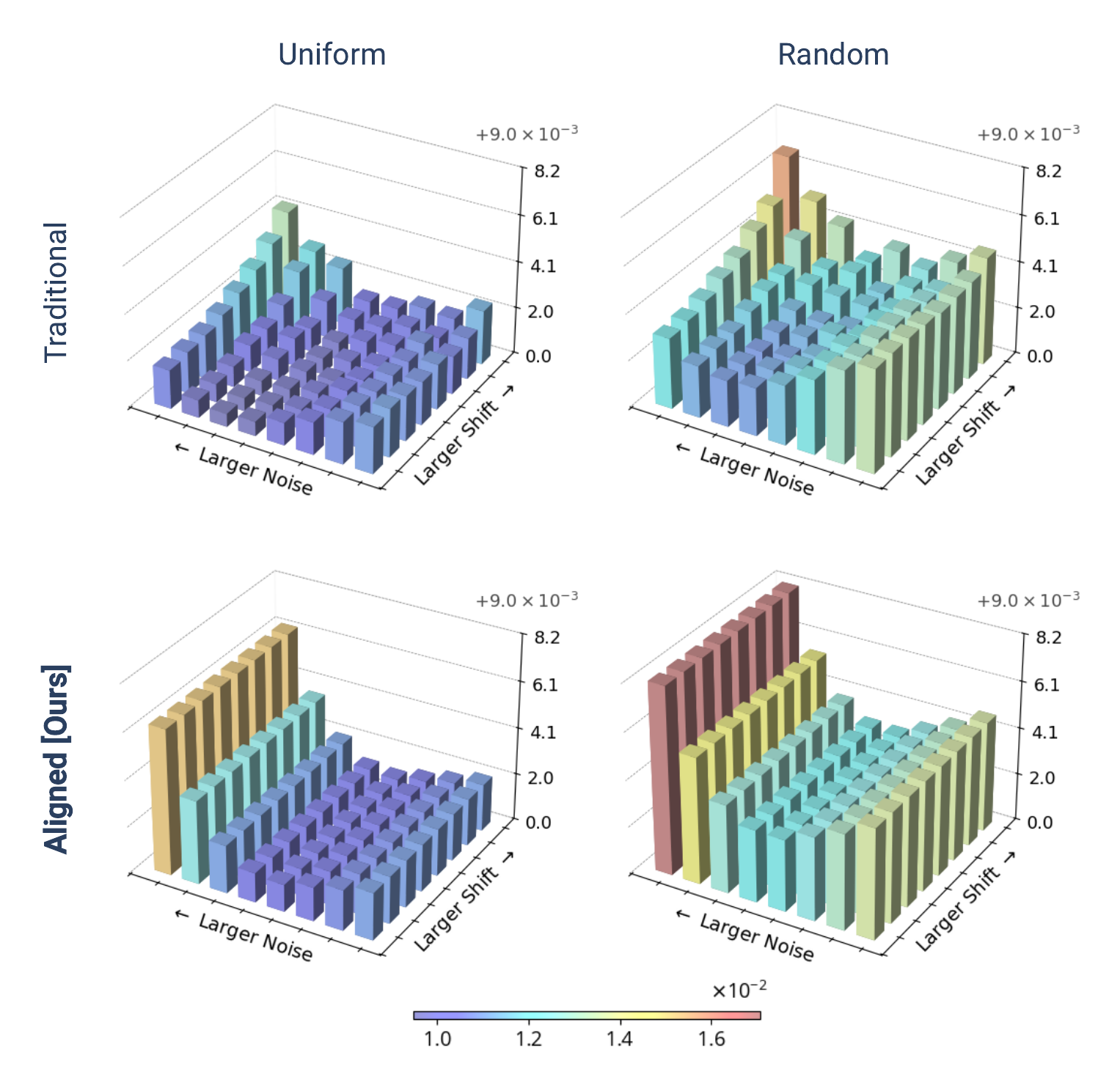}
    \caption{Response of JSD (lower is better) to random noise and barycenter shift on generated samples under various conditions: employing uniform or random sampling of points and with and without the proposed barycenter alignment, i.e. Aligned and Traditional, respectively. }
    \label{fig:JSD}
\end{figure}
\begin{figure*}
    \centering
    \includegraphics[width=\linewidth]{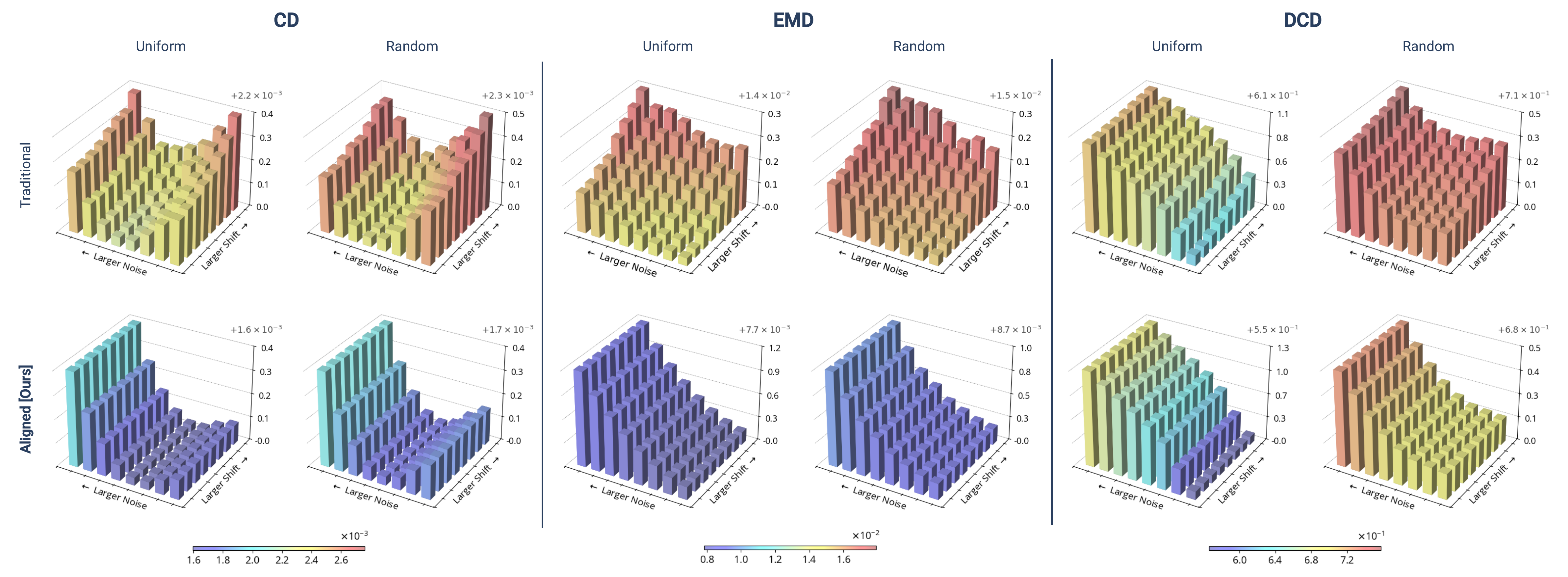}
    \caption{Detailed analysis of the robustness of the MMD metric (lower is better) against noise and sample shifts. The response using different distance measures, CD, EMD, and DCD, to perturbations is shown under four different scenarios: with and without sample barycenter alignment (Aligned and Traditional, respectively) and for Uniform and Random sampling of points, representing uniform and in-homogeneous point distributions.}
    \label{fig:MMD}
\end{figure*}
\begin{figure*}
    \centering
    \includegraphics[width=\linewidth]{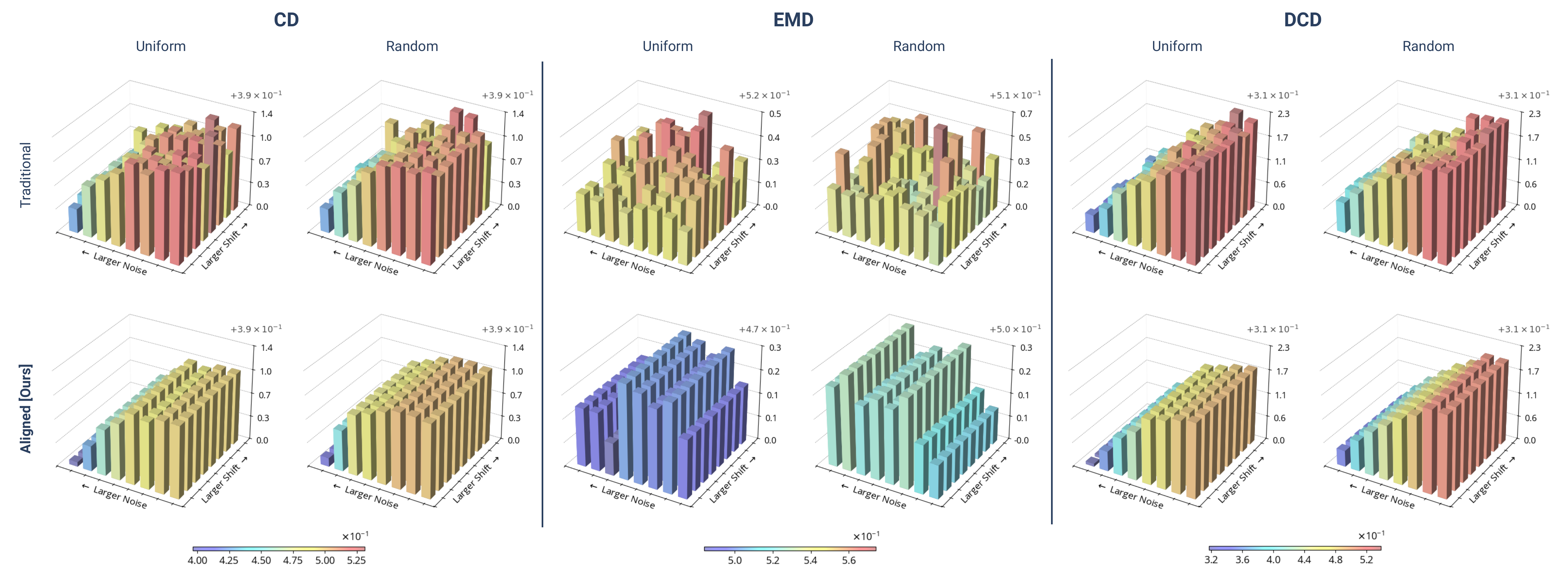}
    \caption{Detailed analysis of the robustness of the COV metric (higher is better) against noise and sample shifts. The response using different distance measures, CD, EMD, and DCD, to perturbations is shown under four different scenarios: with and without sample barycenter alignment (Aligned and Traditional, respectively) and for Uniform and Random sampling of points, representing uniform and in-homogeneous point distributions.}
    \label{fig:COV}
\end{figure*}
\begin{figure*}
    \centering
    \includegraphics[width=\linewidth]{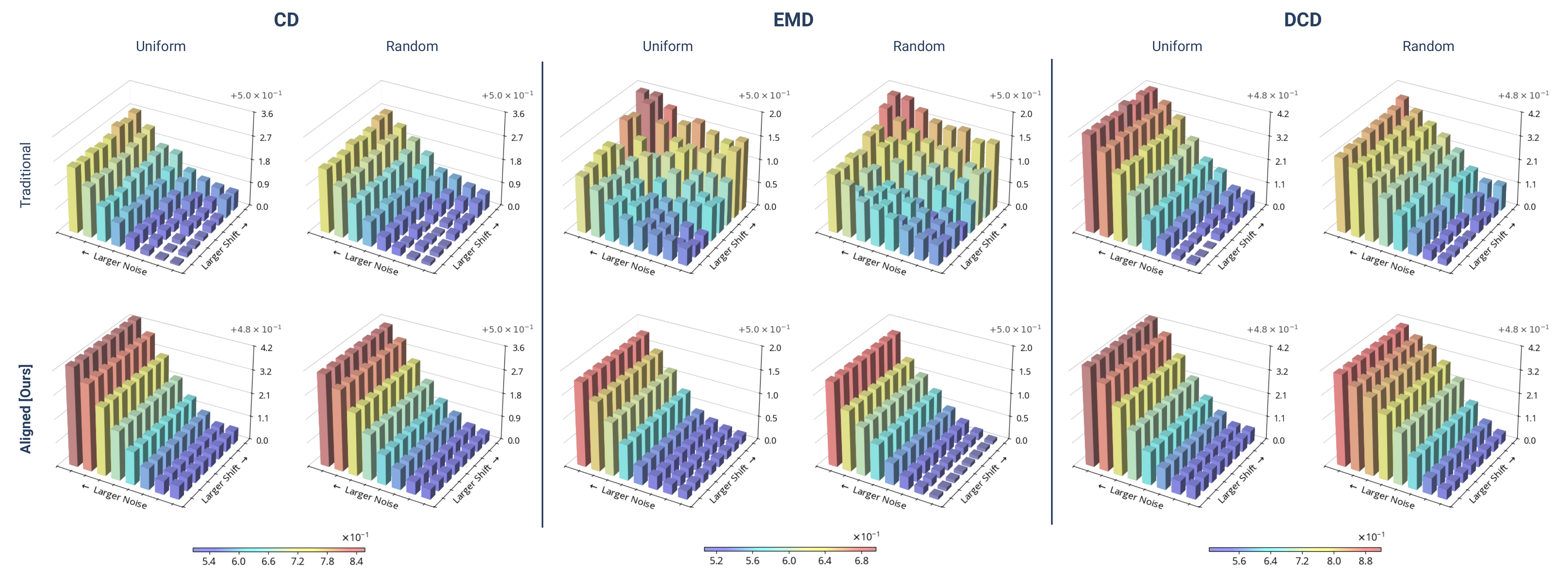}
    \caption{Detailed analysis of the robustness of the 1-NNA metric (lower is better) against noise and sample shifts. The response using different distance measures, CD, EMD, and DCD, to perturbations is shown under four different scenarios: with and without sample barycenter alignment (Aligned and Traditional, respectively) and for Uniform and Random sampling of points, representing uniform and in-homogeneous point distributions.}
    \label{fig:1-NNA}
\end{figure*}
\begin{figure*}
    \centering
    \includegraphics[width=\linewidth]{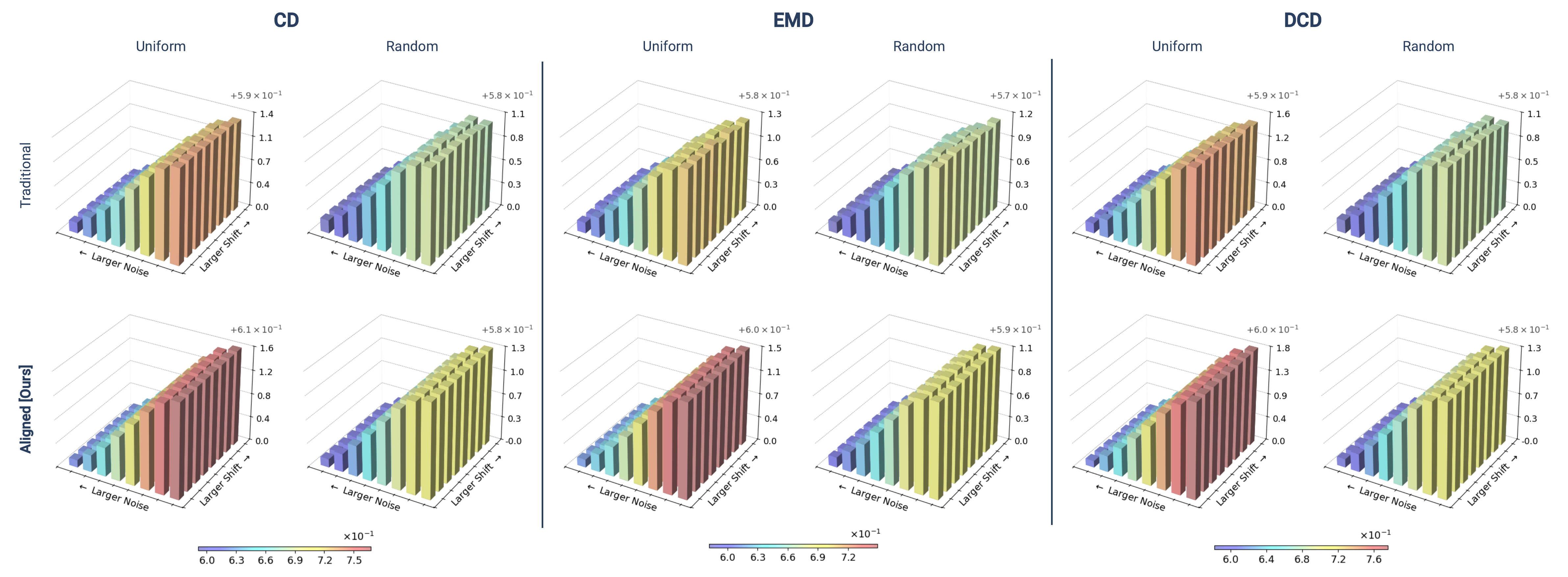}
    \caption{Detailed analysis of the robustness of the proposed SNC metric (higher is better) against noise and sample shifts. The response using different distance measures, CD, EMD, and DCD, to perturbations is shown under four different scenarios: with and without sample barycenter alignment (Aligned and Traditional, respectively) and for Uniform and Random sampling of points, representing uniform and in-homogeneous point distributions.}
    \label{fig:SNC}
\end{figure*}

\textbf{JSD.} Since no distance measure is involved in its calculation, the barycenter alignment proposed in \cref{sec:metrics} for JSD is performed globally rather than pairwise when computing the distance between samples. This means that all the samples are shifted to a common center before assigning each point to a voxel in the voxelized input space and computing the marginal distributions $P_r$ and $P_g$ in \cref{eq:JSD}. In other words, each sample $X \in S_g$ and $Y \in S_r$ is translated to the origin by subtracting its respective barycenter, i.e., $X - \pmb{x}_b$ and $Y - \pmb{y}_b$, where $\pmb{x}_b$ and $\pmb{y}_b$ are the corresponding barycenters. As shown in \cref{fig:JSD}, robustness against shifts is achieved due to global alignment. However, the response of JSD is not monotonic when noise is added to the samples. Consequently, slightly noisy samples may result in a better metric score, failing to provide a reliable assessment of the quality of generated samples. Since JSD is based on marginal probability distributions rather than raw point distances, small amounts of noise can sometimes spread points more uniformly across voxels instead of drastically shifting their distributions. This can make the generated and reference distributions appear more similar, leading to a lower (better) JSD score, even though the actual quality of the samples has degraded. However, at higher noise levels, the distributions become significantly distorted, causing JSD to increase as expected. Given these limitations, JSD is excluded from the comparisons in \cref{tab:model_comparison} and \cref{tab:model_comparison_random} and should be avoided in future evaluations of generative point cloud models.

\begin{table*}[t]
\caption{Comparison of metrics across different models for 3D point cloud generation. All models are evaluated using the same reference set with randomly sampled point clouds and either the set of generated samples published by the authors or pre-trained models when available. MMD is omitted for models trained with per-sample input normalization instead of global normalization, as it does not provide a fair comparison. The best scores are highlighted in bold, while the second best scores are underlined. MMD-DCD is scaled by $10$, and MMD-EMD by $10^3$.}
\label{tab:model_comparison_random}
\footnotesize
\begin{tabularx}{\textwidth}{ll|YY|YY|YY|YY}
\toprule 
 &  \multirow{3}{*}{\vspace{-3.5mm}Model} & \multicolumn{4}{c|}{\textbf{Variability}} & \multicolumn{4}{c}{\textbf{Quality}} \\
 \cmidrule{3-10}
 &  & \multicolumn{2}{c|}{1-NNA (\%,$\downarrow$)} & \multicolumn{2}{c|}{COV (\%,$\uparrow$)} & \multicolumn{2}{c|}{MMD ($\downarrow$)} & \multicolumn{2}{c}{SNC (\%, $\uparrow$)}\\ 
\cmidrule{3-4} \cmidrule{5-6} \cmidrule{7-8} \cmidrule{9-10}
     & & DCD & EMD & DCD & EMD & DCD & EMD & DCD & EMD \\ 
    \midrule

\multirow{9}{*}{\rotatebox[origin=c]{90}{Chair}} & PointFlow \cite{yang_pointflow_2019} & 70.90 & 63.17 & 43.34 & \underline{48.39} & 6.98 & 9.46 & 68.70 & 67.42 \\ 
& SoftFlow \cite{kim_softflow_2020} & 71.13 & 66.92 & 36.60 & 40.74 & 6.97 & 9.56 & 70.75 & 69.06 \\
& ShapeGF \cite{cai_learning_2020} & \underline{57.12} & 65.85 & \textbf{49.46} & 46.09 & - & - & 71.69 & 70.35 \\
& SetVAE \cite{kim_setvae_2021} & 66.16 & 66.39 & 42.27 & 40.89 & \underline{6.94} & 9.29 & \textbf{74.75} & 72.38 \\
& DPM \cite{luo_diffusion_2021} & 80.02 & 91.58 & 40.28 & 31.09 & - & - & 67.91 & 66.59\\
& PVD \cite{zhou_3d_2021} & 58.42 & \textbf{56.05} & 43.34 & 48.09 & 6.96 & 9.12 & 73.15 & 71.42 \\
& LION \cite{zeng_lion_2022} & \textbf{55.51} & \underline{56.74} & \underline{43.80} & \textbf{50.69} & \underline{6.94} & \underline{9.11} & 73.00 & 70.80 \\
& DiT-3D \cite{mo_dit-3d_2023} & 85.60 & 85.99 & 18.07 & 19.45 & 7.13 & 10.42 & 73.50 & 71.76 \\ 
& \textbf{DiPT [Ours]} & 59.65 & 65.16 & 39.82 & 40.89 & \textbf{6.66} & \textbf{9.10} & \underline{74.58} & \textbf{72.76} \\
\midrule
\multirow{9}{*}{\rotatebox[origin=c]{90}{Airplane}} & PointFlow \cite{yang_pointflow_2019} & 86.30 & 75.68 & 39.26 & 43.46 & 5.08 & 2.51 & 81.90 & 79.72 \\
& SoftFlow \cite{kim_softflow_2020} & 81.98 & 69.51 & 43.95 & 47.16 & 4.93 & 2.27 & 82.63 & 81.05 \\
& ShapeGF \cite{cai_learning_2020} & 86.67 & 89.38 & 44.94 & 32.59 & - & - & 81.86 & 79.78 \\
& SetVAE \cite{kim_setvae_2021} & 91.60 & 82.35 & 40.74 & 46.17 & 5.04 & 2.40 & \textbf{85.89} & \textbf{83.79} \\
& DPM \cite{luo_diffusion_2021} & 90.74 & 83.46 & 44.20 & 37.28 & - & - & 81.37 & 79.98 \\
& PVD \cite{zhou_3d_2021} & 82.10 & \underline{67.65} & \underline{45.43} & \textbf{50.12} & 5.11 & \underline{2.26} & 82.92 & 81.15\\
& LION \cite{zeng_lion_2022} & 72.84 & \textbf{65.93} & \textbf{46.17} & \underline{47.41} & \underline{4.95} & \textbf{2.23} &  81.49 & 80.04 \\
& LDT \cite{ji_latent_2024} & \textbf{54.20} & 68.52 & 41.73 & 40.74 & - & - & 84.73 & 82.88 \\ 
&\textbf{DiPT [Ours]} & \underline{62.10} & 87.16 & 36.30 & 37.04 & \textbf{4.54} & 2.53 & \underline{85.20} & \underline{83.16} \\
\midrule
\multirow{8}{*}{\rotatebox[origin=c]{90}{Car}} & PointFlow \cite{yang_pointflow_2019} & 64.81 & 57.26 & 39.89 & 44.16 & 6.17 & 4.61 & 74.61 & 73.05 \\ 
& SoftFlow \cite{kim_softflow_2020} & 66.67 & 63.25 & 34.47 & 41.60 & 6.16 & 4.64 & 75.55 & 73.42 \\
& ShapeGF \cite{cai_learning_2020} & 60.26 & 58.12 &  \underline{48.15} & 43.87 & - & - & 75.17 & 73.26 \\
& SetVAE \cite{kim_setvae_2021} & 65.67 & 66.10 & 35.04 & 37.32 & \underline{6.14} & \underline{4.57} & \textbf{79.65} & \textbf{77.42} \\
& PVD \cite{zhou_3d_2021} & 65.67 & 57.83 & 39.32 & 46.15 & 6.28 & 5.42 & 76.45 & 74.36  \\
& LION \cite{zeng_lion_2022} & 60.26 & \textbf{53.70} & 42.74 & \textbf{51.28} & 6.22 & \textbf{4.52} & 75.55 & 73.46 \\
& LDT \cite{ji_latent_2024} & \textbf{52.56} & \underline{56.70} & \textbf{48.72} & \underline{50.14} & - & - & 79.45 & 76.99 \\ 
&\textbf{DiPT [Ours]} & \underline{54.84} & 73.36 & 29.34 & 32.19 & \textbf{5.77} & 4.83 & \underline{79.23} & \underline{76.83}\\
\bottomrule
\end{tabularx}
\end{table*}

\begin{table}[t]
    \centering
    \caption{Average Spearman correlation of model rankings for each metric, computed relative to uniform and inhomogeneous references, using both DCD and EMD.}

    \footnotesize
    \begin{tabularx}{0.4\textwidth}{l|YYY}
    \toprule
    \multirow{2}{*}{Metric} & \multicolumn{3}{c}{Spearman Correlation} \\
    \cmidrule{2-4}
     & DCD & EMD & Mean \\
    \midrule
    SNC     & $0.96$ & $0.92$ & $0.94$ \\
    MMD     & $0.96$ & $0.30$ & $0.63$ \\
    COV     & $0.75$ & $0.75$ & $0.75$ \\
    1-NNA   & $-0.11$ & $0.39$ & $0.14$ \\
    \bottomrule
    \end{tabularx}
    
    \label{tab:metric_correlations}
\end{table}

\textbf{MMD.} \cref{fig:MMD} shows that the proposed barycenter alignment significantly enhances the stability of MMD. The traditional approach exhibits noticeable, unwanted fluctuations, particularly for CD and EMD. This instability suggests that without explicit alignment, the MMD response becomes unreliable. Furthermore, MMD-CD never exhibits a monotonic increase while adding random perturbations, failing to capture geometric fidelity and local shape consistency, thus providing unreliable qualitative assessment. In contrast, DCD effectively resolves this issue. In fact, MMD-DCD consistently follows a monotonically increasing pattern across both sampling strategies, ensuring a more robust quality evaluation. 

\textbf{COV.} Traditional calculation of COV exhibits strong fluctuations, particularly when computed with EMD, as random shifts are applied to the set of generated samples. This behavior is undesirable since the same samples are always being compared to the references, only in different positions, and should therefore produce always the same COV value. In contrast, the proposed barycenter-aligned approach effectively regularizes the metric, ensuring a consistent response. Moreover, DCD outperforms CD and EMD in preserving a monotonically decreasing metric trend with respect to noise. In comparison, COV-CD remains stable at low noise levels and COV-EMD shows minimal variations across the entire noise range, with a maximum fluctuation of only $2\%$ variation. This behavior can be associated with the nature of EMD. As noise increases, it likely continues to associate noisy samples with the same reference point clouds, since the overall structure of the sample remains unchanged, thus keeping the COV value almost constant. This is an interesting behavior that, along with COV-DCD, can help evaluate the variability of generated samples beside their quality. 

\textbf{1-NNA.} \cref{fig:1-NNA} illustrates the response of the 1-NNA metric to perturbations in the generated samples. The proposed barycenter alignment further enhances an already well-performing metric by stabilizing its value. Moreover, this alignment makes the difference between clean and noisy samples more pronounced when using CD and EMD. 1-NNA exhibits a clear monotonic increase across all distance measures while maintaining similar value ranges for both uniform and in-homogeneous point clouds, particularly for EMD and DCD. 

\begin{figure*}[t]
    \centering
    \includegraphics[width=\linewidth]{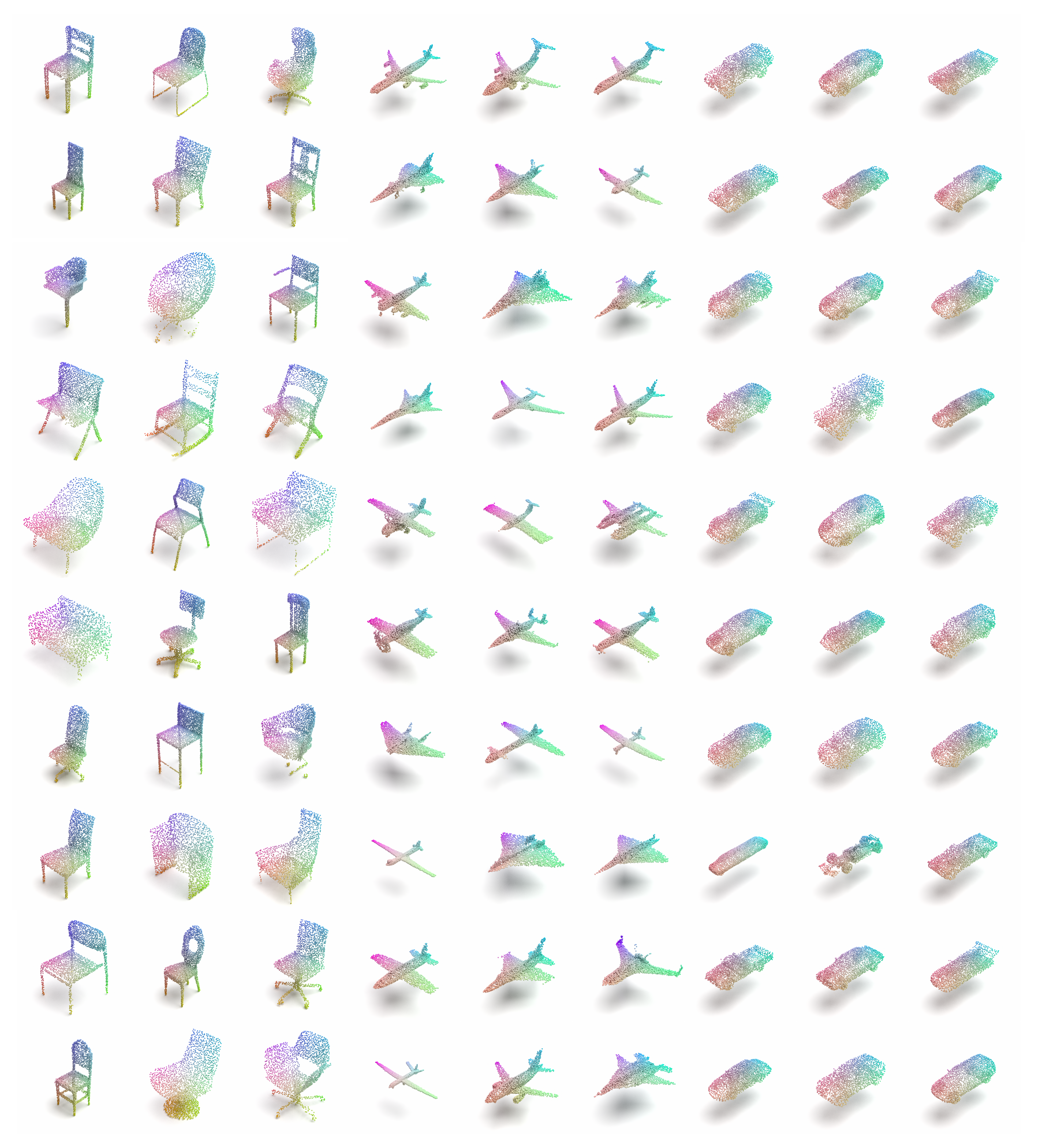}
    \caption{High-fidelity samples randomly generated with the proposed DiPT-S model for chair, airplane and car classes.}
    \label{fig:samples}
\end{figure*}

\begin{table*}[t]
\caption{Quantitative result of the proposed DiPT-S model trained simultaneously on 10 categories for 3D point cloud generation. All categories are evaluated using a generated set of the same size as the reference set with point cloud uniformly sampled. MMD-DCD is scaled by $10$, and MMD-EMD by $10^3$.}
\label{tab:metrics_10cat}
\footnotesize
\begin{tabularx}{\textwidth}{l|YY|YY|YY|YY}
\toprule 
 \multirow{3}{*}{\vspace{-3.5mm}Category} & \multicolumn{4}{c|}{\textbf{Variability}} & \multicolumn{4}{c}{\textbf{Quality}} \\
 \cmidrule{2-9}
 & \multicolumn{2}{c|}{1-NNA (\%,$\downarrow$)} & \multicolumn{2}{c|}{COV (\%,$\uparrow$)} & \multicolumn{2}{c|}{MMD ($\downarrow$)} & \multicolumn{2}{c}{SNC (\%, $\uparrow$)}\\ 
\cmidrule{2-3} \cmidrule{4-5} \cmidrule{6-7} \cmidrule{8-9}
 & DCD & EMD & DCD & EMD & DCD & EMD & DCD & EMD \\ 
    \midrule
    Bathtub & 69.41 & 69.41 & 40.00 & 36.47 & 5.96 & 7.84 & 85.53 & 81.79 \\
    Cap & 50.00 & 60.00 & 40.00 & 40.00 & 7.11 & 12.90 & 77.72 & 77.72 \\
    Bottle & 61.63 & 48.84 & 44.19 & 51.16 & 4.98 & 5.55 & 93.37 & 93.21 \\
    Guitar & 56.25 & 58.13 & 32.50 & 50.00 & 4.36 & 4.23 & 84.55 & 81.63 \\
    Knife & 81.40 & 59.30 & 23.26 & 48.84 & 5.13 & 4.99 & 72.92 & 64.86 \\
    Motorcycle & 79.41 & 60.29 & 41.18 & 41.18 & 5.67 & 5.82 & 63.08 & 62.07  \\
    Mug & 63.64 & 61.36 & 45.45 & 40.91 & 5.65 & 6.08 & 82.85 & 80.37 \\
    Skateboard & 670.00 & 50.00 & 40.00 & 60.00 & 5.18 & 7.90 & 88.06 & 84.69 \\
    Train & 55.13 & 64.10 & 46.15 & 48.72 & 5.23 & 4.81 & 77.48 & 72.52 \\
    Trash Bin & 60.29 & 52.94 & 41.18 & 50.00 & 6.00 & 6.91 & 85.64 & 85.43 \\
\bottomrule
\end{tabularx}
\end{table*}

\begin{table}[t]
\centering
\caption{Number of training and reference samples for each category used of the ShapeNet dataset \cite{chang_shapenet_2015}.}
\label{tab:train_ref}
\footnotesize
\begin{tabularx}{0.3\textwidth}{l|YY}
    \toprule 
    Category &  Training & Reference \\
    \midrule
    Chair & 4612 & 653 \\
    Airplane & 2832 & 405 \\
    Car & 2458 & 351 \\
    \midrule
    Bathtub & 599 & 85 \\
    Cap & 39 & 5 \\
    Bottle & 340 & 43 \\
    Guitar & 557 & 80 \\
    Knife & 296 & 43 \\
    Motorcycle & 235 & 34 \\
    Mug & 149 & 22 \\
    Skateboard & 106 & 15 \\
    Train & 272 & 39 \\
    Trash Bin & 227 & 34 \\
    \bottomrule
\end{tabularx}
\end{table}

\begin{table}[t]
\centering
\caption{Details of DiPT models. We follow ViT \cite{dosovitskiy_image_2021} and DiT \cite{peebles_scalable_2023} model configurations for the Small (S), Base (B) and Large (L) variants. We also introduce an Extra-Small (XS) variant as our smallest model with only $8$ DiPT blocks.}
\label{tab:model_sizes}
\footnotesize
\begin{tabularx}{0.4\textwidth}{l|YYY}
    \toprule 
    Model &  Blocks & Heads & Hidden Size \\
    \midrule
    Extra-Small (XS) & 8 & 6 & 384 \\
    Small (S) & 12 & 6 & 384 \\
    Big (B) & 12 & 12 & 768 \\
    Large (L) & 24 & 16 & 1024 \\ 
    \bottomrule
\end{tabularx}
\end{table}

\begin{figure*}
    \centering
    \includegraphics[width=0.8\linewidth]{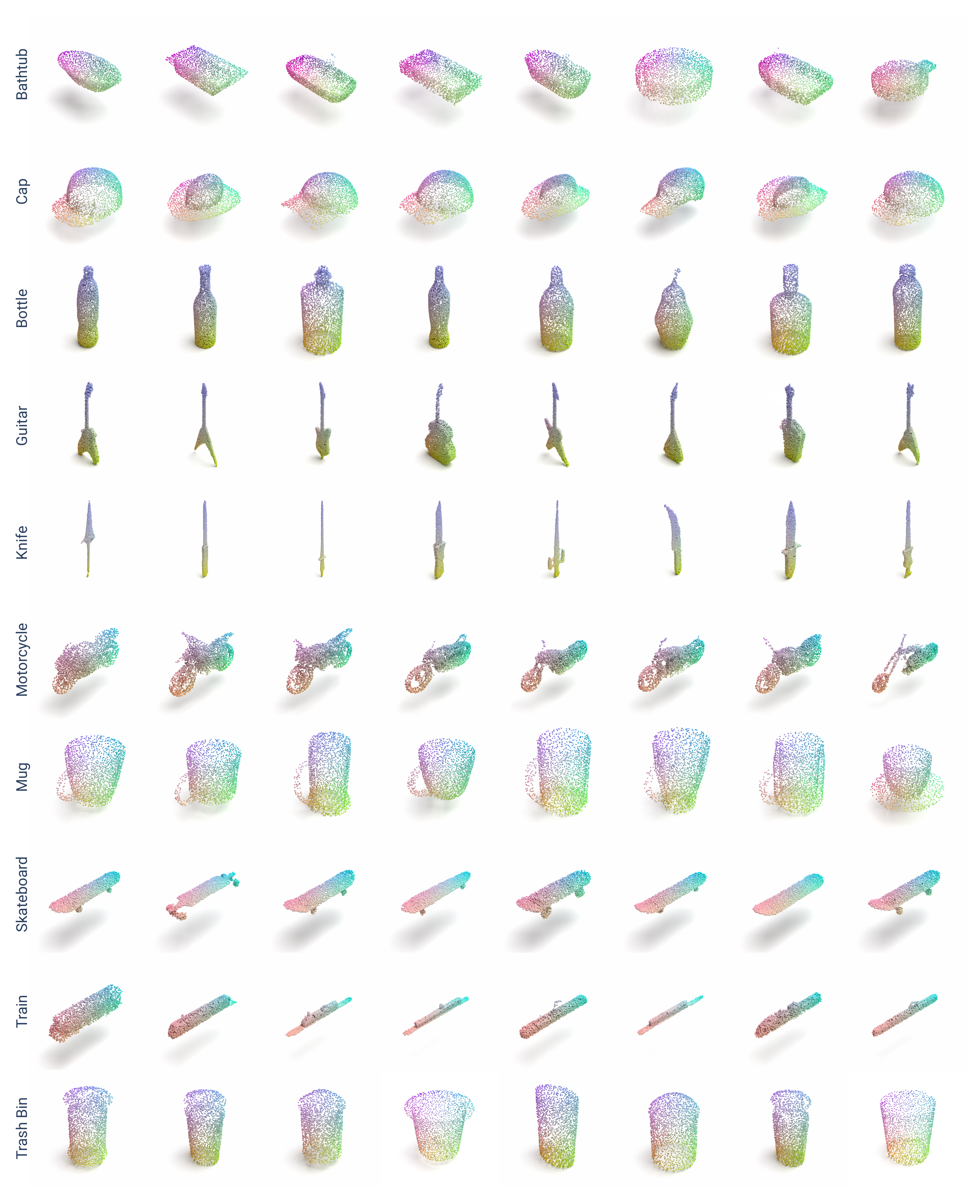}
    \caption{High-fidelity samples randomly generated with the proposed DiPT-S model trained simultaneously on $10$ categories.}
    \label{fig:samples_10cat}
\end{figure*}

\textbf{SNC.} The proposed SNC metric consistently exhibits the desired strong inverse monotonic response to increasing noise, as shown in \cref{fig:SNC}. As previously highlighted in \cref{fig:SNC_param}, there is a significant difference between SNC values computed on uniform and in-homogeneous samples. Once again, this discrepancy arises because normal estimation in sparse regions is less precise, leading to harder point-wise matching. Therefore, to ensure a fair comparison when using SNC to evaluate different generative methods, it is crucial to always use the same reference point clouds, sampled in a consistent manner. To account for this, we provide two separate comparisons in \cref{tab:model_comparison} and \cref{tab:model_comparison_random}, where the same generated samples are compared to uniform and in-homogeneous reference point clouds, respectively.

\section{Additional Experimental Analyses}
\label{sec:additional_experiments}
In this section, we provide additional and more detailed analysis of the experimental results for the proposed DiPT model, complementing \cref{sec:results}.

\textbf{Additional Comparisons with SOTA.} In \cref{tab:model_comparison_random}, we present the same performance comparison of \cref{tab:model_comparison}, but using in-homogeneous point clouds for the reference set, i.e. using random sampling rather than uniform sampling. The same sets of generated samples are used for the metrics calculation as in \cref{tab:model_comparison} and no additional training is conducted. Similar conclusions to those in \cref{sec:results} can be drawn, with our DiPT model outperforming the other methods in terms of the quality of the generated samples. Interestingly, DCD-based metrics show greater consistency across the two comparisons compared to EMD-based metrics. Specifically, the best models according to DCD-based metrics when compared to uniform point clouds are almost always the best when compared to in-homogeneous point clouds as well. In contrast, EMD-based metrics exhibit greater instability, providing sometimes discordant results. This is because EMD, as defined in \cref{eq:EMD}, seeks to minimize the effort required to map a point cloud $X$ into $Y$, and its values can therefore vary depending on the point distribution. Furthermore, SNC also remains largely consistent in the two comparisons, demonstrating its expressive power as a metric. In fact, the average Spearman correlations over categories between metrics computed on uniform and in-homogeneous point clouds, reported in \cref{tab:metric_correlations}, show that SNC-DCD and SNC-EMD have the highest correlations with $0.94$ and $0.92$, respectively, indicating that the metric is robust to reference sampling variations. In contrast, MMD-EMD has a very low correlation of $0.30$, while MMD-DCD has a high correlation of $0.96$. COV shows a moderate correlation of $0.75$ for both DCD and EMD, while 1-NNA has very low correlation for both DCD and EMD.

Therefore, these observations further justify the introduction of DCD-based metrics and SNC for evaluating point cloud generative models. 

\begin{table*}[t]
\caption{Quantitative result of the proposed DiPT model trained with different model sizes for the chair category. The best scores are highlighted in bold. MMD-DCD is scaled by $10$, and MMD-EMD by $10^3$.}
\label{tab:metrics_size}
\footnotesize
\begin{tabularx}{\textwidth}{l|YY|YY|YY|YY|c}
\toprule 
 \multirow{3}{*}{\vspace{-3.5mm}Model} &\multicolumn{4}{c|}{\textbf{Variability}} & \multicolumn{4}{c|}{\textbf{Quality}} & \multirow{3}{*}{\vspace{-3.5mm}Training Time (h)} \\
 \cmidrule{2-9}
 & \multicolumn{2}{c|}{1-NNA (\%,$\downarrow$)} & \multicolumn{2}{c|}{COV (\%,$\uparrow$)} & \multicolumn{2}{c|}{MMD ($\downarrow$)} & \multicolumn{2}{c|}{SNC (\%, $\uparrow$)} &\\ 
\cmidrule{2-3} \cmidrule{4-5} \cmidrule{6-7} \cmidrule{8-9}
 & DCD & EMD & DCD & EMD & DCD & EMD & DCD & EMD & \\ 
    \midrule
    DiPT-XS & 68.91 & 67.30 & 35.99 & 40.74 & 6.11 & 8.75 & 75.54 & 73.46 & 7.40 \\
    DiPT-S & 68.68 & 64.47 & 41.81 & 43.95 & 6.08 & 8.47 & \textbf{77.29} & \textbf{75.10} & 10.40 \\
    DiPT-B & 67.84 & 64.47 & 42.27 & 44.10 & 5.98 & 8.49 & 77.53 & 74.75 & 24.12 \\
    DiPT-L & \textbf{65.62} & \textbf{61.41} & \textbf{48.39} & \textbf{49.00} & \textbf{5.89} & \textbf{8.18} & 76.66 & 74.03 & 70.45 \\
\bottomrule
\end{tabularx}
\end{table*}

\begin{figure*}
    \centering
    \includegraphics[width=0.9\linewidth]{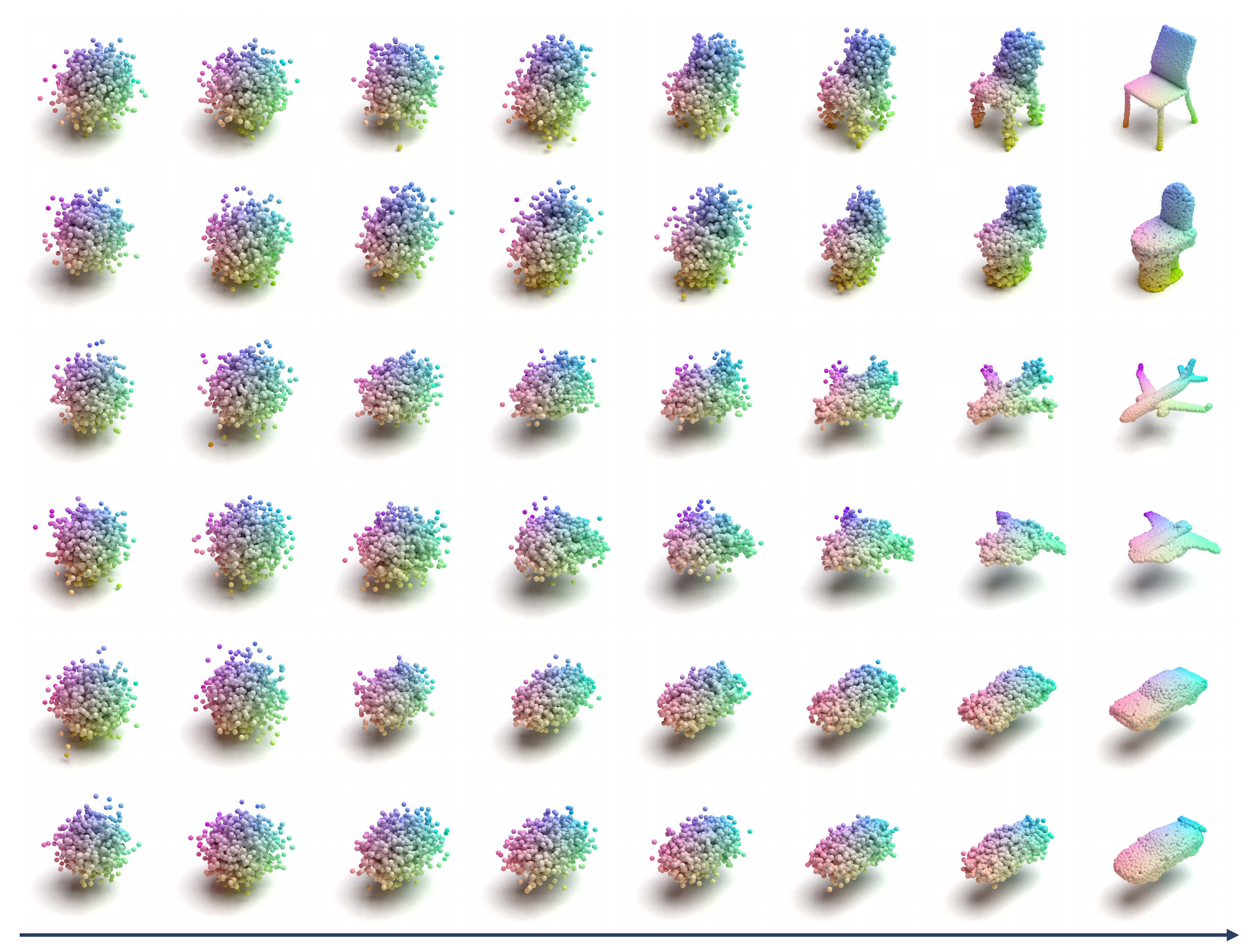}
    \caption{Qualitative visualization of the diffusion process for the chair, airplane, and car categories. The input, i.e., random noise, is shown on the left, while the evolution toward the final shape is displayed moving to the right.}
    \label{fig:evo}
\end{figure*}

\textbf{More Visualizations of Generated Shapes.} To highlight the high-fidelity and diversity of the generated 3D point clouds, we show additional samples from all three categories, chair, airplane and car, generated with DiPT-S in \cref{fig:samples}. These visualizations demonstrate how the proposed model architecture leads to the generation of diverse and high-quality samples for each class, covering a broad spectrum of possible shapes within a given category.

\textbf{Results on 10-Category Training.} To further evaluate the ability of our model to handle multi-class generation under different input conditions, we train DiPT simultaneously on 10 different categories from ShapeNet \cite{chang_shapenet_2015}. Specifically, we use the following classes: bathtub, cap, bottle, guitar, knife, motorcycle, mug, skateboard, train, and trash bin. In \cref{tab:train_ref}, we report the number of samples available in each category for training and testing. Notably, in this experiment, the training samples are significantly fewer than those used in previous experiments with the chair, airplane, and car classes, ranging from a minimum of 39 samples (cap) to a maximum of 599 samples (bathtub). We trained the model using the same settings as in \cref{sec:settings}. \cref{tab:metrics_10cat} presents the quantitative results obtained from evaluating randomly generated samples using the trained model. Moreover, \cref{fig:samples_10cat} illustrates visual examples of these generated samples. Impressively, despite the limited training data, the generated 3D shapes maintain high-quality across all categories. The objects are well-defined and distinct, without noticeable mixing between categories, while also preserving good variability. For instance, even in the two classes with the fewest samples, cap and mug, we observe high-fidelity and diversity, both quantitatively and qualitatively.

\textbf{Effect of Model Size.} We tested the DiPT model following different model sizes as in ViT \cite{dosovitskiy_image_2021} and DiT \cite{peebles_scalable_2023}. Namely, we used the Small (S), Big (B) and Large (L) model sizes. Additionally, we introduced the Extra-Small (XS) size to further shrink the model. \cref{tab:model_sizes} summarizes the differences between different model sizes in terms of number of blocks, attention heads and feature size. The performance comparison between different DiPT sizes is reported in \cref{tab:model_sizes}, focusing only on the chair category for simplicity. Increasing the model size generally improves both the variability and quality of the generated point clouds, whit the Large (L) model bringing significant improvements with respect to the others. Nevertheless, the training time (and consequently, inference time) drastically increases with model size. Therefore, the latter should be selected based on the trade-off between generation quality requirements and available resources.

\textbf{Components Ablation.} Detailed model components ablation (serialization, xCPE) were done in PTv3 \cite{wu_point_2024}. in our additional ablation experiment, downgrading xCPE to RPE resulted in a mean drop of 1.32\% in SNC, 3.26\% in 1-NNA, and 2.42× slower training, showing then the benefit of using xCPE to replace RPE.

\textbf{Evolution of Diffusion Process.} The diffusion process is illustrated for some generated samples from the chair, airplane, and car categories in \cref{fig:evo}. Starting from random noise, the 3D point cloud shapes gradually take form as the diffusion process progresses, ultimately generating a high-fidelity sample in the final steps. Early denoising steps push the points toward the desired shape in an abstract manner, while later steps refine the details, enhancing quality. This common pattern suggests that the initial steps drive the diversity of generated samples, while the final steps are responsible for refining their fidelity and detail.

%% file: diffusion.bib
@article{ji_latent_2024,
	title = {Latent diffusion transformer for point cloud generation},
	volume = {40},
	issn = {1432-2315},
	url = {https://doi.org/10.1007/s00371-024-03396-1},
	doi = {10.1007/s00371-024-03396-1},
	language = {en},
	number = {6},
	urldate = {2024-08-23},
	journal = {The Visual Computer},
	author = {Ji, Junzhong and Zhao, Runfeng and Lei, Minglong},
	month = jun,
	year = {2024},
	pages = {3903--3917},
}

@inproceedings{luo_diffusion_2021,
	address = {Nashville, TN, USA},
	title = {Diffusion {Probabilistic} {Models} for {3D} {Point} {Cloud} {Generation}},
	copyright = {https://ieeexplore.ieee.org/Xplorehelp/downloads/license-information/IEEE.html},
	isbn = {978-1-66544-509-2},
	url = {https://ieeexplore.ieee.org/document/9578791/},
	doi = {10.1109/CVPR46437.2021.00286},
	language = {en},
	urldate = {2024-08-23},
	booktitle = {2021 {IEEE}/{CVF} {Conference} on {Computer} {Vision} and {Pattern} {Recognition} ({CVPR})},
	publisher = {IEEE},
	author = {Luo, Shitong and Hu, Wei},
	month = jun,
	year = {2021},
	pages = {2836--2844},
}

@article{wang_octformer_2023,
	title = {{OctFormer}: {Octree}-based {Transformers} for {3D} {Point} {Clouds}},
	volume = {42},
	issn = {0730-0301, 1557-7368},
	shorttitle = {{OctFormer}},
	url = {http://arxiv.org/abs/2305.03045},
	doi = {10.1145/3592131},
	number = {4},
	urldate = {2024-08-23},
	journal = {ACM Transactions on Graphics},
	author = {Wang, Peng-Shuai},
	month = aug,
	year = {2023},
	note = {arXiv:2305.03045 [cs]},
	pages = {1--11},
}

@article{guo_pct_2021,
	title = {{PCT}: {Point} cloud transformer},
	volume = {7},
	issn = {2096-0433, 2096-0662},
	shorttitle = {{PCT}},
	url = {http://arxiv.org/abs/2012.09688},
	doi = {10.1007/s41095-021-0229-5},
	language = {en},
	number = {2},
	urldate = {2024-08-23},
	journal = {Computational Visual Media},
	author = {Guo, Meng-Hao and Cai, Jun-Xiong and Liu, Zheng-Ning and Mu, Tai-Jiang and Martin, Ralph R. and Hu, Shi-Min},
	month = jun,
	year = {2021},
	note = {arXiv:2012.09688 [cs]},
	pages = {187--199},
}

@inproceedings{yu_point-bert_2022,
	address = {New Orleans, LA, USA},
	title = {Point-{BERT}: {Pre}-training {3D} {Point} {Cloud} {Transformers} with {Masked} {Point} {Modeling}},
	copyright = {https://doi.org/10.15223/policy-029},
	isbn = {978-1-66546-946-3},
	shorttitle = {Point-{BERT}},
	url = {https://ieeexplore.ieee.org/document/9880161/},
	doi = {10.1109/CVPR52688.2022.01871},
	language = {en},
	urldate = {2024-08-23},
	booktitle = {2022 {IEEE}/{CVF} {Conference} on {Computer} {Vision} and {Pattern} {Recognition} ({CVPR})},
	publisher = {IEEE},
	author = {Yu, Xumin and Tang, Lulu and Rao, Yongming and Huang, Tiejun and Zhou, Jie and Lu, Jiwen},
	month = jun,
	year = {2022},
	pages = {19291--19300},
}

@misc{xu_empirical_2018,
	title = {An empirical study on evaluation metrics of generative adversarial networks},
	url = {http://arxiv.org/abs/1806.07755},
	doi = {10.48550/arXiv.1806.07755},
	urldate = {2024-10-24},
	publisher = {arXiv},
	author = {Xu, Qiantong and Huang, Gao and Yuan, Yang and Guo, Chuan and Sun, Yu and Wu, Felix and Weinberger, Kilian},
	month = aug,
	year = {2018},
	note = {arXiv:1806.07755},
}

@article{guerrero_pcpnet_2018,
	title = {{PCPNET}: {Learning} {Local} {Shape} {Properties} from {Raw} {Point} {Clouds}},
	volume = {37},
	issn = {0167-7055, 1467-8659},
	shorttitle = {{PCPNET}},
	url = {http://arxiv.org/abs/1710.04954},
	doi = {10.1111/cgf.13343},
	number = {2},
	urldate = {2025-01-14},
	journal = {Computer Graphics Forum},
	author = {Guerrero, Paul and Kleiman, Yanir and Ovsjanikov, Maks and Mitra, Niloy J.},
	month = may,
	year = {2018},
	note = {arXiv:1710.04954 [cs]},
	pages = {75--85},
}

@inproceedings{lin_hyperbolic_2023,
	title = {Hyperbolic {Chamfer} {Distance} for {Point} {Cloud} {Completion}},
	url = {https://ieeexplore.ieee.org/document/10378095},
	doi = {10.1109/ICCV51070.2023.01342},
	urldate = {2025-01-20},
	booktitle = {2023 {IEEE}/{CVF} {International} {Conference} on {Computer} {Vision} ({ICCV})},
	author = {Lin, Fangzhou and Yue, Yun and Hou, Songlin and Yu, Xuechu and Xu, Yajun and Yamada, Kazunori D and Zhang, Ziming},
	month = oct,
	year = {2023},
	note = {ISSN: 2380-7504},
	pages = {14549--14560},
}

@inproceedings{kim_softflow_2020,
	title = {{SoftFlow}: {Probabilistic} {Framework} for {Normalizing} {Flow} on {Manifolds}},
	volume = {33},
	shorttitle = {{SoftFlow}},
	url = {https://proceedings.neurips.cc/paper/2020/hash/bdbca288fee7f92f2bfa9f7012727740-Abstract.html},
	urldate = {2025-01-21},
	booktitle = {Advances in {Neural} {Information} {Processing} {Systems}},
	publisher = {Curran Associates, Inc.},
	author = {Kim, Hyeongju and Lee, Hyeonseung and Kang, Woo Hyun and Lee, Joun Yeop and Kim, Nam Soo},
	year = {2020},
	pages = {16388--16397},
}

@inproceedings{liu_point-voxel_2019,
	title = {Point-{Voxel} {CNN} for {Efficient} {3D} {Deep} {Learning}},
	volume = {32},
	url = {https://proceedings.neurips.cc/paper_files/paper/2019/hash/5737034557ef5b8c02c0e46513b98f90-Abstract.html},
	urldate = {2025-01-22},
	booktitle = {Advances in {Neural} {Information} {Processing} {Systems}},
	publisher = {Curran Associates, Inc.},
	author = {Liu, Zhijian and Tang, Haotian and Lin, Yujun and Han, Song},
	year = {2019},
}

@inproceedings{liu_flatformer_2023,
	address = {Vancouver, BC, Canada},
	title = {{FlatFormer}: {Flattened} {Window} {Attention} for {Efficient} {Point} {Cloud} {Transformer}},
	copyright = {https://doi.org/10.15223/policy-029},
	isbn = {9798350301298},
	shorttitle = {{FlatFormer}},
	url = {https://ieeexplore.ieee.org/document/10203864/},
	doi = {10.1109/CVPR52729.2023.00122},
	language = {en},
	urldate = {2025-01-22},
	booktitle = {2023 {IEEE}/{CVF} {Conference} on {Computer} {Vision} and {Pattern} {Recognition} ({CVPR})},
	publisher = {IEEE},
	author = {Liu, Zhijian and Yang, Xinyu and Tang, Haotian and Yang, Shang and Han, Song},
	month = jun,
	year = {2023},
	pages = {1200--1211},
}

@misc{wang_serialized_2024,
	title = {Serialized {Point} {Mamba}: {A} {Serialized} {Point} {Cloud} {Mamba} {Segmentation} {Model}},
	shorttitle = {Serialized {Point} {Mamba}},
	url = {http://arxiv.org/abs/2407.12319},
	doi = {10.48550/arXiv.2407.12319},
	urldate = {2025-01-22},
	publisher = {arXiv},
	author = {Wang, Tao and Wen, Wei and Zhai, Jingzhi and Xu, Kang and Luo, Haoming},
	month = jul,
	year = {2024},
	note = {arXiv:2407.12319 [cs]},
}

@book{hilbert_dritter_1935,
	address = {Berlin, Heidelberg},
	title = {Dritter {Band}: {Analysis} · {Grundlagen} der {Mathematik} · {Physik} {Verschiedenes}},
	copyright = {http://www.springer.com/tdm},
	isbn = {978-3-662-37657-7 978-3-662-38452-7},
	shorttitle = {Dritter {Band}},
	url = {http://link.springer.com/10.1007/978-3-662-38452-7},
	language = {de},
	urldate = {2025-01-22},
	publisher = {Springer},
	author = {Hilbert, David},
	year = {1935},
	doi = {10.1007/978-3-662-38452-7},
}

@book{morton_computer_1966,
	title = {A {Computer} {Oriented} {Geodetic} {Data} {Base} and a {New} {Technique} in {File} {Sequencing}},
	language = {en},
	publisher = {International Business Machines Company},
	author = {Morton, G. M.},
	year = {1966},
	note = {Google-Books-ID: 9FFdHAAACAAJ},
}

@inproceedings{liu_swin_2021,
	address = {Montreal, QC, Canada},
	title = {Swin {Transformer}: {Hierarchical} {Vision} {Transformer} using {Shifted} {Windows}},
	copyright = {https://doi.org/10.15223/policy-029},
	isbn = {978-1-66542-812-5},
	shorttitle = {Swin {Transformer}},
	url = {https://ieeexplore.ieee.org/document/9710580/},
	doi = {10.1109/ICCV48922.2021.00986},
	language = {en},
	urldate = {2025-01-22},
	booktitle = {2021 {IEEE}/{CVF} {International} {Conference} on {Computer} {Vision} ({ICCV})},
	publisher = {IEEE},
	author = {Liu, Ze and Lin, Yutong and Cao, Yue and Hu, Han and Wei, Yixuan and Zhang, Zheng and Lin, Stephen and Guo, Baining},
	month = oct,
	year = {2021},
	pages = {9992--10002},
}

@misc{chang_shapenet_2015,
	title = {{ShapeNet}: {An} {Information}-{Rich} {3D} {Model} {Repository}},
	shorttitle = {{ShapeNet}},
	url = {http://arxiv.org/abs/1512.03012},
	doi = {10.48550/arXiv.1512.03012},
	urldate = {2025-01-23},
	publisher = {arXiv},
	author = {Chang, Angel X. and Funkhouser, Thomas and Guibas, Leonidas and Hanrahan, Pat and Huang, Qixing and Li, Zimo and Savarese, Silvio and Savva, Manolis and Song, Shuran and Su, Hao and Xiao, Jianxiong and Yi, Li and Yu, Fisher},
	month = dec,
	year = {2015},
	note = {arXiv:1512.03012 [cs]},
}

@inproceedings{myronenko_non-rigid_2006,
	title = {Non-rigid point set registration: {Coherent} {Point} {Drift}},
	volume = {19},
	shorttitle = {Non-rigid point set registration},
	url = {https://proceedings.neurips.cc/paper_files/paper/2006/hash/3b2d8f129ae2f408f2153cd9ce663043-Abstract.html},
	urldate = {2025-01-31},
	booktitle = {Advances in {Neural} {Information} {Processing} {Systems}},
	publisher = {MIT Press},
	author = {Myronenko, Andriy and Song, Xubo and Carreira-Perpiñán, Miguel},
	year = {2006},
}

@misc{yu_3d_2021,
	title = {{3D} {Medical} {Point} {Transformer}: {Introducing} {Convolution} to {Attention} {Networks} for {Medical} {Point} {Cloud} {Analysis}},
	shorttitle = {{3D} {Medical} {Point} {Transformer}},
	url = {http://arxiv.org/abs/2112.04863},
	doi = {10.48550/arXiv.2112.04863},
	urldate = {2025-02-03},
	publisher = {arXiv},
	author = {Yu, Jianhui and Zhang, Chaoyi and Wang, Heng and Zhang, Dingxin and Song, Yang and Xiang, Tiange and Liu, Dongnan and Cai, Weidong},
	month = dec,
	year = {2021},
	note = {arXiv:2112.04863 [eess]},
}

@article{duan_robotics_2021,
	title = {Robotics {Dexterous} {Grasping}: {The} {Methods} {Based} on {Point} {Cloud} and {Deep} {Learning}},
	volume = {15},
	issn = {1662-5218},
	shorttitle = {Robotics {Dexterous} {Grasping}},
	url = {https://www.frontiersin.org/journals/neurorobotics/articles/10.3389/fnbot.2021.658280/full},
	doi = {10.3389/fnbot.2021.658280},
	language = {English},
	urldate = {2025-02-03},
	journal = {Frontiers in Neurorobotics},
	author = {Duan, Haonan and Wang, Peng and Huang, Yayu and Xu, Guangyun and Wei, Wei and Shen, Xiaofei},
	month = jun,
	year = {2021},
	note = {Publisher: Frontiers},
}

@misc{ramesh_hierarchical_2022,
	title = {Hierarchical {Text}-{Conditional} {Image} {Generation} with {CLIP} {Latents}},
	url = {http://arxiv.org/abs/2204.06125},
	doi = {10.48550/arXiv.2204.06125},
	urldate = {2025-02-03},
	publisher = {arXiv},
	author = {Ramesh, Aditya and Dhariwal, Prafulla and Nichol, Alex and Chu, Casey and Chen, Mark},
	month = apr,
	year = {2022},
	note = {arXiv:2204.06125 [cs]},
}

@inproceedings{ran_surface_2022,
	title = {Surface {Representation} for {Point} {Clouds}},
	url = {https://ieeexplore.ieee.org/document/9880155},
	doi = {10.1109/CVPR52688.2022.01837},
	urldate = {2025-02-04},
	booktitle = {2022 {IEEE}/{CVF} {Conference} on {Computer} {Vision} and {Pattern} {Recognition} ({CVPR})},
	author = {Ran, Haoxi and Liu, Jun and Wang, Chengjie},
	month = jun,
	year = {2022},
	note = {ISSN: 2575-7075},
	pages = {18920--18930},
}

@article{huang_surface_2024,
	title = {Surface {Reconstruction} {From} {Point} {Clouds}: {A} {Survey} and a {Benchmark}},
	volume = {46},
	issn = {1939-3539},
	shorttitle = {Surface {Reconstruction} {From} {Point} {Clouds}},
	url = {https://ieeexplore.ieee.org/document/10599623},
	doi = {10.1109/TPAMI.2024.3429209},
	number = {12},
	urldate = {2025-02-04},
	journal = {IEEE Transactions on Pattern Analysis and Machine Intelligence},
	author = {Huang, ZhangJin and Wen, Yuxin and Wang, ZiHao and Ren, Jinjuan and Jia, Kui},
	month = dec,
	year = {2024},
	note = {Conference Name: IEEE Transactions on Pattern Analysis and Machine Intelligence},
	pages = {9727--9748},
}

@article{liu_grab-net_2023,
	title = {{GRAB}-{Net}: {Graph}-{Based} {Boundary}-{Aware} {Network} for {Medical} {Point} {Cloud} {Segmentation}},
	volume = {42},
	issn = {1558-254X},
	shorttitle = {{GRAB}-{Net}},
	url = {https://ieeexplore.ieee.org/document/10093984/?arnumber=10093984},
	doi = {10.1109/TMI.2023.3265000},
	number = {9},
	urldate = {2025-02-04},
	journal = {IEEE Transactions on Medical Imaging},
	author = {Liu, Yifan and Li, Wuyang and Liu, Jie and Chen, Hui and Yuan, Yixuan},
	month = sep,
	year = {2023},
	note = {Conference Name: IEEE Transactions on Medical Imaging},
	pages = {2776--2786},
}

@article{triess_realism_2022,
	title = {A {Realism} {Metric} for {Generated} {LiDAR} {Point} {Clouds}},
	volume = {130},
	issn = {1573-1405},
	url = {https://doi.org/10.1007/s11263-022-01676-8},
	doi = {10.1007/s11263-022-01676-8},
	language = {en},
	number = {12},
	urldate = {2025-02-05},
	journal = {International Journal of Computer Vision},
	author = {Triess, Larissa T. and Rist, Christoph B. and Peter, David and Zöllner, J. Marius},
	month = dec,
	year = {2022},
	pages = {2962--2979},
}

@inproceedings{ho_denoising_2020,
	address = {Red Hook, NY, USA},
	series = {{NIPS} '20},
	title = {Denoising diffusion probabilistic models},
	isbn = {978-1-71382-954-6},
	urldate = {2025-02-05},
	booktitle = {Proceedings of the 34th {International} {Conference} on {Neural} {Information} {Processing} {Systems}},
	publisher = {Curran Associates Inc.},
	author = {Ho, Jonathan and Jain, Ajay and Abbeel, Pieter},
	year = {2020},
	pages = {6840--6851},
}

@article{ho_video_2022,
	title = {Video {Diffusion} {Models}},
	volume = {35},
	url = {https://proceedings.neurips.cc/paper_files/paper/2022/hash/39235c56aef13fb05a6adc95eb9d8d66-Abstract-Conference.html},
	language = {en},
	urldate = {2025-02-06},
	journal = {Advances in Neural Information Processing Systems},
	author = {Ho, Jonathan and Salimans, Tim and Gritsenko, Alexey and Chan, William and Norouzi, Mohammad and Fleet, David J.},
	month = dec,
	year = {2022},
	pages = {8633--8646},
}

@inproceedings{huang_fastdiff_2022,
	address = {Vienna, Austria},
	title = {{FastDiff}: {A} {Fast} {Conditional} {Diffusion} {Model} for {High}-{Quality} {Speech} {Synthesis}},
	isbn = {978-1-956792-00-3},
	shorttitle = {{FastDiff}},
	url = {https://www.ijcai.org/proceedings/2022/577},
	doi = {10.24963/ijcai.2022/577},
	language = {en},
	urldate = {2025-02-06},
	booktitle = {Proceedings of the {Thirty}-{First} {International} {Joint} {Conference} on {Artificial} {Intelligence}},
	publisher = {International Joint Conferences on Artificial Intelligence Organization},
	author = {Huang, Rongjie and Lam, Max W. Y. and Wang, Jun and Su, Dan and Yu, Dong and Ren, Yi and Zhao, Zhou},
	month = jul,
	year = {2022},
	pages = {4157--4163},
}

@inproceedings{lee_set_2019,
	title = {Set {Transformer}: {A} {Framework} for {Attention}-based {Permutation}-{Invariant} {Neural} {Networks}},
	shorttitle = {Set {Transformer}},
	url = {https://proceedings.mlr.press/v97/lee19d.html},
	language = {en},
	urldate = {2025-02-17},
	booktitle = {Proceedings of the 36th {International} {Conference} on {Machine} {Learning}},
	publisher = {PMLR},
	author = {Lee, Juho and Lee, Yoonho and Kim, Jungtaek and Kosiorek, Adam and Choi, Seungjin and Teh, Yee Whye},
	month = may,
	year = {2019},
	note = {ISSN: 2640-3498},
	pages = {3744--3753},
}

@article{kullback_information_1951,
	title = {On {Information} and {Sufficiency}},
	volume = {22},
	issn = {0003-4851, 2168-8990},
	url = {https://projecteuclid.org/journals/annals-of-mathematical-statistics/volume-22/issue-1/On-Information-and-Sufficiency/10.1214/aoms/1177729694.full},
	doi = {10.1214/aoms/1177729694},
	number = {1},
	urldate = {2025-02-17},
	journal = {The Annals of Mathematical Statistics},
	author = {Kullback, S. and Leibler, R. A.},
	month = mar,
	year = {1951},
	note = {Publisher: Institute of Mathematical Statistics},
	pages = {79--86},
}

@article{berkmann_computation_1994,
	title = {Computation of surface geometry and segmentation using covariance techniques},
	volume = {16},
	issn = {1939-3539},
	url = {https://ieeexplore.ieee.org/document/334391},
	doi = {10.1109/34.334391},
	number = {11},
	urldate = {2025-02-18},
	journal = {IEEE Transactions on Pattern Analysis and Machine Intelligence},
	author = {Berkmann, J. and Caelli, T.},
	month = nov,
	year = {1994},
	note = {Conference Name: IEEE Transactions on Pattern Analysis and Machine Intelligence},
	pages = {1114--1116},
}

@inproceedings{klasing_comparison_2009,
	title = {Comparison of surface normal estimation methods for range sensing applications},
	url = {https://ieeexplore.ieee.org/document/5152493},
	doi = {10.1109/ROBOT.2009.5152493},
	urldate = {2025-02-18},
	booktitle = {2009 {IEEE} {International} {Conference} on {Robotics} and {Automation}},
	author = {Klasing, Klaas and Althoff, Daniel and Wollherr, Dirk and Buss, Martin},
	month = may,
	year = {2009},
	note = {ISSN: 1050-4729},
	pages = {3206--3211},
}

@article{hoppe_surface_1992,
	title = {Surface reconstruction from unorganized points},
	volume = {26},
	issn = {0097-8930},
	url = {https://dl.acm.org/doi/10.1145/142920.134011},
	doi = {10.1145/142920.134011},
	number = {2},
	urldate = {2025-02-18},
	journal = {SIGGRAPH Comput. Graph.},
	author = {Hoppe, Hugues and DeRose, Tony and Duchamp, Tom and McDonald, John and Stuetzle, Werner},
	year = {1992},
	pages = {71--78},
}

@inproceedings{behley_semantickitti_2019,
	address = {Seoul, Korea (South)},
	title = {{SemanticKITTI}: {A} {Dataset} for {Semantic} {Scene} {Understanding} of {LiDAR} {Sequences}},
	copyright = {https://ieeexplore.ieee.org/Xplorehelp/downloads/license-information/IEEE.html},
	isbn = {978-1-72814-803-8},
	shorttitle = {{SemanticKITTI}},
	url = {https://ieeexplore.ieee.org/document/9010727/},
	doi = {10.1109/ICCV.2019.00939},
	language = {en},
	urldate = {2025-02-19},
	booktitle = {2019 {IEEE}/{CVF} {International} {Conference} on {Computer} {Vision} ({ICCV})},
	publisher = {IEEE},
	author = {Behley, Jens and Garbade, Martin and Milioto, Andres and Quenzel, Jan and Behnke, Sven and Stachniss, Cyrill and Gall, Jurgen},
	month = oct,
	year = {2019},
	pages = {9296--9306},
}

@article{chen_3d_2021,
	title = {{3D} {Point} {Cloud} {Processing} and {Learning} for {Autonomous} {Driving}: {Impacting} {Map} {Creation}, {Localization}, and {Perception}},
	volume = {38},
	issn = {1558-0792},
	shorttitle = {{3D} {Point} {Cloud} {Processing} and {Learning} for {Autonomous} {Driving}},
	url = {https://ieeexplore.ieee.org/document/9307334},
	doi = {10.1109/MSP.2020.2984780},
	number = {1},
	urldate = {2025-02-19},
	journal = {IEEE Signal Processing Magazine},
	author = {Chen, Siheng and Liu, Baoan and Feng, Chen and Vallespi-Gonzalez, Carlos and Wellington, Carl},
	month = jan,
	year = {2021},
	note = {Conference Name: IEEE Signal Processing Magazine},
	pages = {68--86},
}

@article{dao_flashattention_2022,
	title = {{FlashAttention}: {Fast} and {Memory}-{Efficient} {Exact} {Attention} with {IO}-{Awareness}},
	volume = {35},
	shorttitle = {{FlashAttention}},
	url = {https://proceedings.neurips.cc/paper_files/paper/2022/hash/67d57c32e20fd0a7a302cb81d36e40d5-Abstract-Conference.html},
	language = {en},
	urldate = {2025-02-19},
	journal = {Advances in Neural Information Processing Systems},
	author = {Dao, Tri and Fu, Dan and Ermon, Stefano and Rudra, Atri and Ré, Christopher},
	month = dec,
	year = {2022},
	pages = {16344--16359},
}

@inproceedings{dhariwal_diffusion_2021,
	title = {Diffusion {Models} {Beat} {GANs} on {Image} {Synthesis}},
	volume = {34},
	url = {https://proceedings.nips.cc/paper/2021/hash/49ad23d1ec9fa4bd8d77d02681df5cfa-Abstract.html},
	urldate = {2025-02-19},
	booktitle = {Advances in {Neural} {Information} {Processing} {Systems}},
	publisher = {Curran Associates, Inc.},
	author = {Dhariwal, Prafulla and Nichol, Alexander},
	year = {2021},
	pages = {8780--8794},
}

@inproceedings{gadelha_multiresolution_2018,
	address = {Berlin, Heidelberg},
	title = {Multiresolution {Tree} {Networks} for {3D} {Point} {Cloud} {Processing}},
	isbn = {978-3-030-01233-5},
	url = {https://doi.org/10.1007/978-3-030-01234-2_7},
	doi = {10.1007/978-3-030-01234-2_7},
	urldate = {2025-02-19},
	booktitle = {Computer {Vision} – {ECCV} 2018: 15th {European} {Conference}, {Munich}, {Germany}, {September} 8–14, 2018, {Proceedings}, {Part} {VII}},
	publisher = {Springer-Verlag},
	author = {Gadelha, Matheus and Wang, Rui and Maji, Subhransu},
	year = {2018},
	pages = {105--122},
}

@inproceedings{heusel_gans_2017,
	address = {Red Hook, NY, USA},
	series = {{NIPS}'17},
	title = {{GANs} trained by a two time-scale update rule converge to a local nash equilibrium},
	isbn = {978-1-5108-6096-4},
	urldate = {2025-02-19},
	booktitle = {Proceedings of the 31st {International} {Conference} on {Neural} {Information} {Processing} {Systems}},
	publisher = {Curran Associates Inc.},
	author = {Heusel, Martin and Ramsauer, Hubert and Unterthiner, Thomas and Nessler, Bernhard and Hochreiter, Sepp},
	year = {2017},
	pages = {6629--6640},
}

@inproceedings{kim_setvae_2021,
	address = {Nashville, TN, USA},
	title = {{SetVAE}: {Learning} {Hierarchical} {Composition} for {Generative} {Modeling} of {Set}-{Structured} {Data}},
	copyright = {https://ieeexplore.ieee.org/Xplorehelp/downloads/license-information/IEEE.html},
	isbn = {978-1-66544-509-2},
	shorttitle = {{SetVAE}},
	url = {https://ieeexplore.ieee.org/document/9577267/},
	doi = {10.1109/CVPR46437.2021.01481},
	language = {en},
	urldate = {2025-02-19},
	booktitle = {2021 {IEEE}/{CVF} {Conference} on {Computer} {Vision} and {Pattern} {Recognition} ({CVPR})},
	publisher = {IEEE},
	author = {Kim, Jinwoo and Yoo, Jaehoon and Lee, Juho and Hong, Seunghoon},
	month = jun,
	year = {2021},
	pages = {15054--15063},
}

@article{liu_self-supervised_2023,
	title = {Self-{Supervised} {Point} {Cloud} {Registration} {With} {Deep} {Versatile} {Descriptors} for {Intelligent} {Driving}},
	volume = {24},
	issn = {1558-0016},
	url = {https://ieeexplore.ieee.org/document/10109135},
	doi = {10.1109/TITS.2023.3268273},
	number = {9},
	urldate = {2025-02-19},
	journal = {IEEE Transactions on Intelligent Transportation Systems},
	author = {Liu, Dongrui and Chen, Chuanchaun and Xu, Changqing and Qiu, Robert C. and Chu, Lei},
	month = sep,
	year = {2023},
	note = {Conference Name: IEEE Transactions on Intelligent Transportation Systems},
	pages = {9767--9779},
}

@inproceedings{mo_dit-3d_2023,
	address = {Red Hook, NY, USA},
	series = {{NIPS} '23},
	title = {{DiT}-{3D}: exploring plain diffusion transformers for {3D} shape generation},
	shorttitle = {{DiT}-{3D}},
	urldate = {2025-02-19},
	booktitle = {Proceedings of the 37th {International} {Conference} on {Neural} {Information} {Processing} {Systems}},
	publisher = {Curran Associates Inc.},
	author = {Mo, Shentong and Xie, Enze and Chu, Ruihang and Yao, Lewei and Hong, Lanqing and Nießner, Matthias and Li, Zhenguo},
	year = {2023},
	pages = {67960--67971},
}

@inproceedings{nguyen_point-set_2021,
	address = {Montreal, QC, Canada},
	title = {Point-set {Distances} for {Learning} {Representations} of {3D} {Point} {Clouds}},
	copyright = {https://doi.org/10.15223/policy-029},
	isbn = {978-1-66542-812-5},
	url = {https://ieeexplore.ieee.org/document/9711238/},
	doi = {10.1109/ICCV48922.2021.01031},
	language = {en},
	urldate = {2025-02-19},
	booktitle = {2021 {IEEE}/{CVF} {International} {Conference} on {Computer} {Vision} ({ICCV})},
	publisher = {IEEE},
	author = {Nguyen, Trung and Pham, Quang-Hieu and Le, Tam and Pham, Tung and Ho, Nhat and Hua, Binh-Son},
	month = oct,
	year = {2021},
	pages = {10458--10467},
}

@inproceedings{petrov_gem3d_2024,
	address = {New York, NY, USA},
	series = {{SIGGRAPH} '24},
	title = {{GEM3D}: {GEnerative} {Medial} {Abstractions} for {3D} {Shape} {Synthesis}},
	isbn = {9798400705250},
	shorttitle = {{GEM3D}},
	url = {https://dl.acm.org/doi/10.1145/3641519.3657415},
	doi = {10.1145/3641519.3657415},
	urldate = {2025-02-19},
	booktitle = {{ACM} {SIGGRAPH} 2024 {Conference} {Papers}},
	publisher = {Association for Computing Machinery},
	author = {Petrov, Dmitry and Goyal, Pradyumn and Thamizharasan, Vikas and Kim, Vladimir and Gadelha, Matheus and Averkiou, Melinos and Chaudhuri, Siddhartha and Kalogerakis, Evangelos},
	year = {2024},
	pages = {1--11},
}

@inproceedings{qi_pointnet_2017,
	title = {{PointNet}++: {Deep} {Hierarchical} {Feature} {Learning} on {Point} {Sets} in a {Metric} {Space}},
	volume = {30},
	shorttitle = {{PointNet}++},
	url = {https://proceedings.neurips.cc/paper_files/paper/2017/hash/d8bf84be3800d12f74d8b05e9b89836f-Abstract.html},
	urldate = {2025-02-19},
	booktitle = {Advances in {Neural} {Information} {Processing} {Systems}},
	publisher = {Curran Associates, Inc.},
	author = {Qi, Charles Ruizhongtai and Yi, Li and Su, Hao and Guibas, Leonidas J},
	year = {2017},
}

@article{shah_flashattention-3_2025,
	title = {{FlashAttention}-3: {Fast} and {Accurate} {Attention} with {Asynchrony} and {Low}-precision},
	volume = {37},
	shorttitle = {{FlashAttention}-3},
	url = {https://proceedings.neurips.cc/paper_files/paper/2024/hash/7ede97c3e082c6df10a8d6103a2eebd2-Abstract-Conference.html},
	language = {en},
	urldate = {2025-02-19},
	journal = {Advances in Neural Information Processing Systems},
	author = {Shah, Jay and Bikshandi, Ganesh and Zhang, Ying and Thakkar, Vijay and Ramani, Pradeep and Dao, Tri},
	month = jan,
	year = {2025},
	pages = {68658--68685},
}

@inproceedings{shu_3d_2019,
	address = {Seoul, Korea (South)},
	title = {{3D} {Point} {Cloud} {Generative} {Adversarial} {Network} {Based} on {Tree} {Structured} {Graph} {Convolutions}},
	copyright = {https://ieeexplore.ieee.org/Xplorehelp/downloads/license-information/IEEE.html},
	isbn = {978-1-72814-803-8},
	url = {https://ieeexplore.ieee.org/document/9009495/},
	doi = {10.1109/ICCV.2019.00396},
	language = {en},
	urldate = {2025-02-19},
	booktitle = {2019 {IEEE}/{CVF} {International} {Conference} on {Computer} {Vision} ({ICCV})},
	publisher = {IEEE},
	author = {Shu, Dongwook and Park, Sung Woo and Kwon, Junseok},
	month = oct,
	year = {2019},
	pages = {3858--3867},
}

@inproceedings{vaswani_attention_2017,
	title = {Attention is {All} you {Need}},
	volume = {30},
	url = {https://proceedings.neurips.cc/paper_files/paper/2017/hash/3f5ee243547dee91fbd053c1c4a845aa-Abstract.html},
	urldate = {2025-02-19},
	booktitle = {Advances in {Neural} {Information} {Processing} {Systems}},
	publisher = {Curran Associates, Inc.},
	author = {Vaswani, Ashish and Shazeer, Noam and Parmar, Niki and Uszkoreit, Jakob and Jones, Llion and Gomez, Aidan N and Kaiser, Ł ukasz and Polosukhin, Illia},
	year = {2017},
}

@inproceedings{wang_weighted_2023,
	title = {Weighted {Point} {Cloud} {Normal} {Estimation}},
	url = {https://ieeexplore.ieee.org/document/10219584},
	doi = {10.1109/ICME55011.2023.00345},
	urldate = {2025-02-19},
	booktitle = {2023 {IEEE} {International} {Conference} on {Multimedia} and {Expo} ({ICME})},
	author = {Wang, Weijia and Lu, Xuequan and Shao, Di and Liu, Xiao and Dazeley, Richard and Robles-Kelly, Antonio and Pan, Wei},
	month = jul,
	year = {2023},
	note = {ISSN: 1945-788X},
	pages = {2015--2020},
}

@inproceedings{wu_point_2024,
	address = {Seattle, WA, USA},
	title = {Point {Transformer} {V3}: {Simpler}, {Faster}, {Stronger}},
	copyright = {https://doi.org/10.15223/policy-029},
	isbn = {9798350353006},
	shorttitle = {Point {Transformer} {V3}},
	url = {https://ieeexplore.ieee.org/document/10658198/},
	doi = {10.1109/CVPR52733.2024.00463},
	language = {en},
	urldate = {2025-02-19},
	booktitle = {2024 {IEEE}/{CVF} {Conference} on {Computer} {Vision} and {Pattern} {Recognition} ({CVPR})},
	publisher = {IEEE},
	author = {Wu, Xiaoyang and Jiang, Li and Wang, Peng-Shuai and Liu, Zhijian and Liu, Xihui and Qiao, Yu and Ouyang, Wanli and He, Tong and Zhao, Hengshuang},
	month = jun,
	year = {2024},
	pages = {4840--4851},
}

@inproceedings{wu_point_2022,
	address = {Red Hook, NY, USA},
	series = {{NIPS} '22},
	title = {Point transformer {V2}: grouped vector attention and partition-based pooling},
	isbn = {978-1-71387-108-8},
	shorttitle = {Point transformer {V2}},
	urldate = {2025-02-19},
	booktitle = {Proceedings of the 36th {International} {Conference} on {Neural} {Information} {Processing} {Systems}},
	publisher = {Curran Associates Inc.},
	author = {Wu, Xiaoyang and Lao, Yixing and Jiang, Li and Liu, Xihui and Zhao, Hengshuang},
	month = nov,
	year = {2022},
	pages = {33330--33342},
}

@inproceedings{yang_pointflow_2019,
	address = {Seoul, Korea (South)},
	title = {{PointFlow}: {3D} {Point} {Cloud} {Generation} {With} {Continuous} {Normalizing} {Flows}},
	copyright = {https://ieeexplore.ieee.org/Xplorehelp/downloads/license-information/IEEE.html},
	isbn = {978-1-72814-803-8},
	shorttitle = {{PointFlow}},
	url = {https://ieeexplore.ieee.org/document/9010395/},
	doi = {10.1109/ICCV.2019.00464},
	language = {en},
	urldate = {2025-02-19},
	booktitle = {2019 {IEEE}/{CVF} {International} {Conference} on {Computer} {Vision} ({ICCV})},
	publisher = {IEEE},
	author = {Yang, Guandao and Huang, Xun and Hao, Zekun and Liu, Ming-Yu and Belongie, Serge and Hariharan, Bharath},
	month = oct,
	year = {2019},
	pages = {4540--4549},
}

@inproceedings{yang_foldingnet_2018,
	address = {Salt Lake City, UT},
	title = {{FoldingNet}: {Point} {Cloud} {Auto}-{Encoder} via {Deep} {Grid} {Deformation}},
	isbn = {978-1-5386-6420-9},
	shorttitle = {{FoldingNet}},
	url = {https://ieeexplore.ieee.org/document/8578127/},
	doi = {10.1109/CVPR.2018.00029},
	language = {en},
	urldate = {2025-02-19},
	booktitle = {2018 {IEEE}/{CVF} {Conference} on {Computer} {Vision} and {Pattern} {Recognition}},
	publisher = {IEEE},
	author = {Yang, Yaoqing and Feng, Chen and Shen, Yiru and Tian, Dong},
	month = jun,
	year = {2018},
	pages = {206--215},
}

@inproceedings{zeng_lion_2022,
	address = {Red Hook, NY, USA},
	series = {{NIPS} '22},
	title = {{LION}: latent point diffusion models for {3D} shape generation},
	isbn = {978-1-71387-108-8},
	shorttitle = {{LION}},
	urldate = {2025-02-19},
	booktitle = {Proceedings of the 36th {International} {Conference} on {Neural} {Information} {Processing} {Systems}},
	publisher = {Curran Associates Inc.},
	author = {Zeng, Xiaohui and Vahdat, Arash and Williams, Francis and Gojcic, Zan and Litany, Or and Fidler, Sanja and Kreis, Karsten},
	month = nov,
	year = {2022},
	pages = {10021--10039},
}

@inproceedings{zhou_3d_2021,
	address = {Montreal, QC, Canada},
	title = {{3D} {Shape} {Generation} and {Completion} through {Point}-{Voxel} {Diffusion}},
	copyright = {https://doi.org/10.15223/policy-029},
	isbn = {978-1-66542-812-5},
	url = {https://ieeexplore.ieee.org/document/9711332/},
	doi = {10.1109/ICCV48922.2021.00577},
	language = {en},
	urldate = {2025-02-19},
	booktitle = {2021 {IEEE}/{CVF} {International} {Conference} on {Computer} {Vision} ({ICCV})},
	publisher = {IEEE},
	author = {Zhou, Linqi and Du, Yilun and Wu, Jiajun},
	month = oct,
	year = {2021},
	pages = {5806--5815},
}

@article{zhou_fast_2022,
	title = {Fast and {Accurate} {Normal} {Estimation} for {Point} {Clouds} {Via} {Patch} {Stitching}},
	volume = {142},
	issn = {0010-4485},
	url = {https://doi.org/10.1016/j.cad.2021.103121},
	doi = {10.1016/j.cad.2021.103121},
	number = {C},
	urldate = {2025-02-19},
	journal = {Comput. Aided Des.},
	author = {Zhou, Jun and Jin, Wei and Wang, Mingjie and Liu, Xiuping and Li, Zhiyang and Liu, Zhaobin},
	year = {2022},
}

@inproceedings{cai_learning_2020,
	address = {Berlin, Heidelberg},
	title = {Learning {Gradient} {Fields} for {Shape} {Generation}},
	isbn = {978-3-030-58579-2},
	url = {https://doi.org/10.1007/978-3-030-58580-8_22},
	doi = {10.1007/978-3-030-58580-8_22},
	urldate = {2025-02-19},
	booktitle = {Computer {Vision} – {ECCV} 2020: 16th {European} {Conference}, {Glasgow}, {UK}, {August} 23–28, 2020, {Proceedings}, {Part} {III}},
	publisher = {Springer-Verlag},
	author = {Cai, Ruojin and Yang, Guandao and Averbuch-Elor, Hadar and Hao, Zekun and Belongie, Serge and Snavely, Noah and Hariharan, Bharath},
	year = {2020},
	pages = {364--381},
}

@inproceedings{klokov_discrete_2020,
	address = {Berlin, Heidelberg},
	title = {Discrete {Point} {Flow} {Networks} for {Efficient} {Point} {Cloud} {Generation}},
	isbn = {978-3-030-58591-4},
	url = {https://doi.org/10.1007/978-3-030-58592-1_41},
	doi = {10.1007/978-3-030-58592-1_41},
	urldate = {2025-02-19},
	booktitle = {Computer {Vision} – {ECCV} 2020: 16th {European} {Conference}, {Glasgow}, {UK}, {August} 23–28, 2020, {Proceedings}, {Part} {XXIII}},
	publisher = {Springer-Verlag},
	author = {Klokov, Roman and Boyer, Edmond and Verbeek, Jakob},
	year = {2020},
	pages = {694--710},
}

@inproceedings{wu_density-aware_2021,
	address = {Red Hook, NY, USA},
	series = {{NIPS} '21},
	title = {Density-aware chamfer distance as a comprehensive metric for point cloud completion},
	isbn = {978-1-71384-539-3},
	urldate = {2025-02-19},
	booktitle = {Proceedings of the 35th {International} {Conference} on {Neural} {Information} {Processing} {Systems}},
	publisher = {Curran Associates Inc.},
	author = {Wu, Tong and Pan, Liang and Zhang, Junzhe and Wang, Tai and Liu, Ziwei and Lin, Dahua},
	year = {2021},
	pages = {29088--29100},
}

@inproceedings{ben-shabat_nesti-net_2019,
	address = {Long Beach, CA, USA},
	title = {Nesti-{Net}: {Normal} {Estimation} for {Unstructured} {3D} {Point} {Clouds} {Using} {Convolutional} {Neural} {Networks}},
	copyright = {https://ieeexplore.ieee.org/Xplorehelp/downloads/license-information/IEEE.html},
	isbn = {978-1-72813-293-8},
	shorttitle = {Nesti-{Net}},
	url = {https://ieeexplore.ieee.org/document/8953652/},
	doi = {10.1109/CVPR.2019.01035},
	language = {en},
	urldate = {2025-02-19},
	booktitle = {2019 {IEEE}/{CVF} {Conference} on {Computer} {Vision} and {Pattern} {Recognition} ({CVPR})},
	publisher = {IEEE},
	author = {Ben-Shabat, Yizhak and Lindenbaum, Michael and Fischer, Anath},
	month = jun,
	year = {2019},
	pages = {10104--10112},
}

@inproceedings{lai_stratified_2022,
	title = {Stratified {Transformer} for {3D} {Point} {Cloud} {Segmentation}},
	url = {https://ieeexplore.ieee.org/abstract/document/9879705},
	doi = {10.1109/CVPR52688.2022.00831},
	urldate = {2025-02-19},
	booktitle = {2022 {IEEE}/{CVF} {Conference} on {Computer} {Vision} and {Pattern} {Recognition} ({CVPR})},
	author = {Lai, Xin and Liu, Jianhui and Jiang, Li and Wang, Liwei and Zhao, Hengshuang and Liu, Shu and Qi, Xiaojuan and Jia, Jiaya},
	month = jun,
	year = {2022},
	note = {ISSN: 2575-7075},
	pages = {8490--8499},
}

@inproceedings{mitra_estimating_2003,
	address = {New York, NY, USA},
	series = {{SCG} '03},
	title = {Estimating surface normals in noisy point cloud data},
	isbn = {978-1-58113-663-0},
	url = {https://doi.org/10.1145/777792.777840},
	doi = {10.1145/777792.777840},
	urldate = {2025-02-19},
	booktitle = {Proceedings of the nineteenth annual symposium on {Computational} geometry},
	publisher = {Association for Computing Machinery},
	author = {Mitra, Niloy J. and Nguyen, An},
	year = {2003},
	pages = {322--328},
}

@article{pauly_shape_2003,
	title = {Shape modeling with point-sampled geometry},
	volume = {22},
	issn = {0730-0301},
	url = {https://doi.org/10.1145/882262.882319},
	doi = {10.1145/882262.882319},
	number = {3},
	urldate = {2025-02-19},
	journal = {ACM Trans. Graph.},
	author = {Pauly, Mark and Keiser, Richard and Kobbelt, Leif P. and Gross, Markus},
	year = {2003},
	pages = {641--650},
}

@inproceedings{peebles_scalable_2023,
	title = {Scalable {Diffusion} {Models} with {Transformers}},
	url = {https://ieeexplore.ieee.org/document/10377858/?arnumber=10377858},
	doi = {10.1109/ICCV51070.2023.00387},
	urldate = {2025-02-19},
	booktitle = {2023 {IEEE}/{CVF} {International} {Conference} on {Computer} {Vision} ({ICCV})},
	author = {Peebles, William and Xie, Saining},
	month = oct,
	year = {2023},
	note = {ISSN: 2380-7504},
	pages = {4172--4182},
}

@inproceedings{achlioptas_learning_2018,
	title = {Learning {Representations} and {Generative} {Models} for {3D} {Point} {Clouds}},
	url = {https://proceedings.mlr.press/v80/achlioptas18a.html},
	language = {en},
	urldate = {2025-02-19},
	booktitle = {Proceedings of the 35th {International} {Conference} on {Machine} {Learning}},
	publisher = {PMLR},
	author = {Achlioptas, Panos and Diamanti, Olga and Mitliagkas, Ioannis and Guibas, Leonidas},
	month = jul,
	year = {2018},
	note = {ISSN: 2640-3498},
	pages = {40--49},
}

@inproceedings{valsesia_learning_2019,
	title = {Learning {Localized} {Generative} {Models} for {3D} {Point} {Clouds} via {Graph} {Convolution}},
	url = {https://openreview.net/forum?id=SJeXSo09FQ},
	booktitle = {7th {International} {Conference} on {Learning} {Representations}, {ICLR} 2019, {New} {Orleans}, {LA}, {USA}, {May} 6-9, 2019},
	publisher = {OpenReview.net},
	author = {Valsesia, Diego and Fracastoro, Giulia and Magli, Enrico},
	year = {2019},
}

@inproceedings{xia_diffir_2023,
	title = {{DiffIR}: {Efficient} {Diffusion} {Model} for {Image} {Restoration}},
	shorttitle = {{DiffIR}},
	url = {https://ieeexplore.ieee.org/document/10377629/?arnumber=10377629},
	doi = {10.1109/ICCV51070.2023.01204},
	urldate = {2025-02-19},
	booktitle = {2023 {IEEE}/{CVF} {International} {Conference} on {Computer} {Vision} ({ICCV})},
	author = {Xia, Bin and Zhang, Yulun and Wang, Shiyin and Wang, Yitong and Wu, Xinglong and Tian, Yapeng and Yang, Wenming and Van Gool, Luc},
	month = oct,
	year = {2023},
	note = {ISSN: 2380-7504},
	pages = {13049--13059},
}

@inproceedings{zhao_point_2021,
	title = {Point {Transformer}},
	url = {https://ieeexplore.ieee.org/abstract/document/9710703},
	doi = {10.1109/ICCV48922.2021.01595},
	urldate = {2025-02-19},
	booktitle = {2021 {IEEE}/{CVF} {International} {Conference} on {Computer} {Vision} ({ICCV})},
	author = {Zhao, Hengshuang and Jiang, Li and Jia, Jiaya and Torr, Philip and Koltun, Vladlen},
	month = oct,
	year = {2021},
	note = {ISSN: 2380-7504},
	pages = {16239--16248},
}

@inproceedings{lopez-paz_revisiting_2017,
	title = {Revisiting {Classifier} {Two}-{Sample} {Tests}},
	url = {https://openreview.net/forum?id=SJkXfE5xx},
	booktitle = {5th {International} {Conference} on {Learning} {Representations}, {ICLR} 2017, {Toulon}, {France}, {April} 24-26, 2017, {Conference} {Track} {Proceedings}},
	publisher = {OpenReview.net},
	author = {Lopez-Paz, David and Oquab, Maxime},
	year = {2017},
}

@inproceedings{smith_super-convergence_2019,
	title = {Super-convergence: very fast training of neural networks using large learning rates},
	volume = {11006},
	shorttitle = {Super-convergence},
	url = {https://www.spiedigitallibrary.org/conference-proceedings-of-spie/11006/1100612/Super-convergence--very-fast-training-of-neural-networks-using/10.1117/12.2520589.full},
	doi = {10.1117/12.2520589},
	urldate = {2025-02-19},
	booktitle = {Artificial {Intelligence} and {Machine} {Learning} for {Multi}-{Domain} {Operations} {Applications}},
	publisher = {SPIE},
	author = {Smith, Leslie N. and Topin, Nicholay},
	month = may,
	year = {2019},
	pages = {369--386},
}

@inproceedings{loshchilov_decoupled_2019,
	title = {Decoupled {Weight} {Decay} {Regularization}},
	url = {https://openreview.net/forum?id=Bkg6RiCqY7},
	booktitle = {7th {International} {Conference} on {Learning} {Representations}, {ICLR} 2019, {New} {Orleans}, {LA}, {USA}, {May} 6-9, 2019},
	publisher = {OpenReview.net},
	author = {Loshchilov, Ilya and Hutter, Frank},
	year = {2019},
}

@inproceedings{dosovitskiy_image_2021,
	title = {An {Image} is {Worth} 16x16 {Words}: {Transformers} for {Image} {Recognition} at {Scale}},
	url = {https://openreview.net/forum?id=YicbFdNTTy},
	booktitle = {9th {International} {Conference} on {Learning} {Representations}, {ICLR} 2021, {Virtual} {Event}, {Austria}, {May} 3-7, 2021},
	publisher = {OpenReview.net},
	author = {Dosovitskiy, Alexey and Beyer, Lucas and Kolesnikov, Alexander and Weissenborn, Dirk and Zhai, Xiaohua and Unterthiner, Thomas and Dehghani, Mostafa and Minderer, Matthias and Heigold, Georg and Gelly, Sylvain and Uszkoreit, Jakob and Houlsby, Neil},
	year = {2021},
}

@article{ruthotto_introduction_2021,
	title = {An introduction to deep generative modeling},
	volume = {44},
	issn = {1522-2608},
	url = {https://onlinelibrary.wiley.com/doi/10.1002/gamm.202100008},
	doi = {10.1002/gamm.202100008},
	language = {en},
	number = {2},
	urldate = {2025-02-19},
	journal = {GAMM-Mitteilungen},
	author = {Ruthotto, Lars and Haber, Eldad},
	month = jun,
	year = {2021},
	note = {Publisher: John Wiley \& Sons, Ltd},
	pages = {e202100008},
}

@inproceedings{dao_flashattention-2_2024,
	title = {{FlashAttention}-2: {Faster} {Attention} with {Better} {Parallelism} and {Work} {Partitioning}},
	url = {https://openreview.net/forum?id=mZn2Xyh9Ec},
	booktitle = {The {Twelfth} {International} {Conference} on {Learning} {Representations}, {ICLR} 2024, {Vienna}, {Austria}, {May} 7-11, 2024},
	publisher = {OpenReview.net},
	author = {Dao, Tri},
	year = {2024},
}

@inproceedings{jo_graph_2024,
	title = {Graph {Generation} with {Diffusion} {Mixture}},
	url = {https://openreview.net/forum?id=cZTFxktg23},
	booktitle = {Forty-first {International} {Conference} on {Machine} {Learning}, {ICML} 2024, {Vienna}, {Austria}, {July} 21-27, 2024},
	publisher = {OpenReview.net},
	author = {Jo, Jaehyeong and Kim, Dongki and Hwang, Sung Ju},
	year = {2024},
}
